\documentclass[journal]{IEEEtran}

\usepackage{graphicx}
\usepackage{subfigure}
\usepackage{amsmath}
\usepackage{amssymb}
\usepackage{bm}
\usepackage{threeparttable}
\usepackage{multirow}
\usepackage{cases}
\usepackage[ruled,linesnumbered]{algorithm2e}

\usepackage{booktabs}

\newtheorem{remark}{Remark}

\usepackage{pifont}

\ifCLASSINFOpdf
\else
\fi

\begin{document}

\title{AeRSoM: An Aerial Rigid-Soft Integrated Manipulator for Contact-Rich Manipulation}



\author{Jiacheng Liang,
        Hang Zhong, 
        Yaonan Wang,
        Ge Chen,
        Zhixing Zhang,
        Bocheng Tian,
        Hui Zhang, 
        Li Wen
\thanks{Corresponding author: Hang Zhong, e-mail: zhonghang@hnu.edu.cn.}
\thanks{Jiacheng Liang, Hang Zhong, Yaonan Wang, Ge Chen, Zhixing Zhang, and Hui Zhang are with the School of Artificial Intelligence and Robotics, Hunan University, Changsha 410082, China and also with the National Engineering Research Center for Robot Visual Perception and Control Technology, Changsha 410082, China (e-mail: liangjiacheng@hnu.edu.cn; zhonghang@hnu.edu.cn; yaonan@hnu.edu.cn; arcg@hnu.edu.cn; zhangzhixing@hnu.edu.cn; zhanghui1983@hnu.edu.cn).}
\thanks{Bocheng Tian and Li Wen are with the School of Mechanical Engineering and Automation, Beihang University, Beijing 100191, China (e-mail: 19376448@buaa.edu.cn; liwen@buaa.edu.cn).}
}


\maketitle

\begin{abstract}

Contact-rich aerial manipulation remains fundamentally challenging because interaction forces are directly transmitted to the aerial platform, often leading to instability and degraded task performance.
While compliant manipulators can mitigate these effects, existing aerial manipulation systems typically struggle to reconcile interaction compliance with manipulation precision.
To this end, this article presents an aerial rigid-soft integrated manipulator (AeRSoM) robot that realizes embodied compliance for aerial manipulation.
The proposed system integrates a fully actuated aerial platform, a rigid-soft manipulator, and variable-stiffness regulation to simultaneously achieve stable flight, compliant interaction, and precise manipulation.
By distributing compliance throughout the manipulation system, the proposed design leverages distributed embodied compliance to passively absorb contact disturbances while preserving sufficient stiffness for task execution.
To fully exploit the mechanical design, a composite control framework is developed for precise end-effector trajectory tracking in the presence of uncertainties and external disturbances.
Extensive real-world experiments are conducted in representative contact-rich aerial manipulation tasks, including dynamic transmission-line grasping, physical interaction with a wind turbine blade, peg-in-hole, and screwing operations.
The results demonstrate that the proposed rigid-soft integration significantly improves interaction robustness and task adaptability while maintaining manipulation accuracy, highlighting that embodied compliance provides a promising design paradigm for enhancing the safety, robustness, and versatility of aerial manipulation.

\end{abstract}


\IEEEpeerreviewmaketitle

\section{Introduction}

\IEEEPARstart{A}{erial} manipulation has emerged as a promising capability for performing inspection, maintenance, and intervention tasks in environments that are difficult or hazardous for human workers \cite{2025ZhongTASE,2022OlleroTRO,2022ZhangNature}.
Despite significant progress in aerial robotic platforms and manipulation systems, enabling reliable physical interaction with the environment remains a fundamental challenge.
Unlike free-flight operations, contact-rich aerial manipulation introduces interaction forces that are directly transmitted to the aerial platform, potentially causing instability, trajectory deviations, and task failures.
As a result, achieving robust physical interaction while maintaining precise manipulation continues to be one of the central challenges in the aerial robotics community.

Most existing aerial manipulators rely on rigid robotic arms to achieve accurate positioning and force transmission.
Although rigid structures provide manipulation precision and payload capability, their high structural stiffness limits the system's ability to adapt to contact impacts and environmental errors, making interaction-sensitive tasks particularly challenging.
Consequently, successful operation often requires accurate environment models, sophisticated force regulation strategies, and precise state estimation \cite{2019RyllIJRR,2021BodieTRO}.
These requirements become increasingly difficult to satisfy in unstructured environments characterized by uncertainties, disturbances, and imperfect sensing.

Compliance has recently gained traction as a key mechanism for enhancing the safety and robustness of aerial manipulation \cite{2024MelletRoboSoft}.
Through passive deformation during contact, compliant structures can absorb impact energy, redistribute interaction forces, and enlarge task tolerances.
More importantly, compliance enables part of the interaction complexity to be handled through physical morphology rather than active control, thereby reducing the burden on sensing and feedback regulation.
However, introducing compliance typically comes at the cost of reduced stiffness, lower manipulation precision, and limited load-bearing capability, creating a fundamental tradeoff between robust interaction and accurate manipulation.
Moreover, while compliant components have been integrated into rigid links to enhance compliance, these solutions frequently fall short in terms of dexterity and maneuverability.

Several studies have explored the integration of soft robotic components into aerial manipulation systems \cite{2022SzaszRoboSoft,2022JalaliFRAI}.
While these approaches improve interaction safety and environmental adaptability, they often suffer from insufficient structural stiffness, reduced manipulation accuracy, or restricted payload capability.
Furthermore, many existing systems rely on fixed compliance characteristics, limiting their ability to accommodate the diverse requirements of different manipulation stages.
Therefore, achieving a balanced combination of compliance, precision, and adaptability remains an open problem in aerial manipulation.

To address these challenges, this article proposes an aerial rigid-soft integrated manipulator (AeRSoM) capable of embodied compliant aerial manipulation.
The core concept is to leverage distributed compliance within the system while maintaining sufficient structural rigidity for precise task execution.
The proposed system combines a fully actuated aerial base paired with a rigid-soft integrated manipulation.
Inspired by the flexibility and maneuverability of human hands, this manipulator features a rigid rotary mechanism combined with a lightweight soft robotic arm, expanding the operational range and enhancing compliance in aerial tasks, as shown in Fig. \ref{fig_scenario}.
Additionally, the chain-mail jamming technique is employed to provide the soft arm with variable stiffness, enabling rigidity-compliance adjustments during different stages of manipulation.
Through this rigid-soft integration, the system can passively attenuate disturbances upon contact, adapt to environmental uncertainties, and maintain accuracy across a broad range of aerial manipulation tasks.
The main contributions of this work are summarized as follows:

\begin{figure}
	\centering
	\includegraphics[width=3.3in]{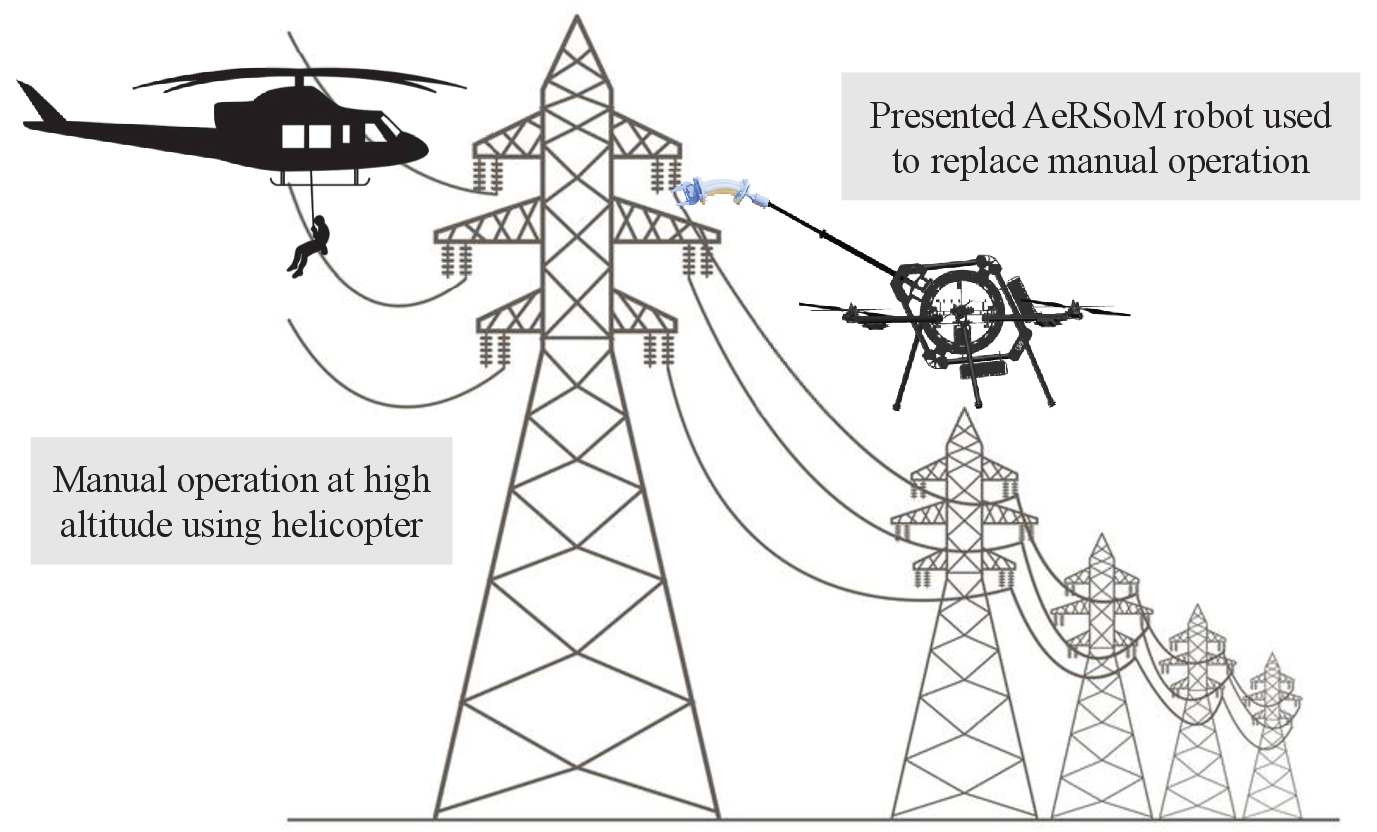}
	\caption{Schematic diagram of manual operation at high altitude using helicopter (left), and using the presented AeRSoM robot to replace human for aerial manipulation (right).}
	\label{fig_scenario}
\end{figure}

\begin{itemize}
	\item We propose a rigid-soft integrated aerial manipulation paradigm that enables embodied compliant interaction through distributed compliance and variable stiffness regulation.
	This paradigm enables a unique combination of fully-actuated flight, enlarged manipulation workspace, and distributed compliance, which is difficult to achieve simultaneously by existing aerial manipulators.
	\item We develop a modeling and control framework that addresses actuator hysteresis, payload variations, and environmental disturbances to achieve robust and precise aerial manipulation.
	\item Comprehensive real-world experiments are conducted in representative contact-rich aerial manipulation tasks and provide promising results.
	In particular, comparative experiments involving multi-directional dynamic transmission-line grasping are performed using the presented robot equipped with a soft arm versus one with a purely rigid arm, highlighting the advantages of compliance without the requirements for additional complex and expensive interaction methods.
	Furthermore, the flexibility and performance of the end-effector are showcased through physical interaction with a wind turbine blade featuring an unstructured surface.
	Finally, peg-in-hole and screwing experiments are conducted to further validate the capabilities of the presented robot for complex aerial manipulation tasks.
	\item Experimental results demonstrate that embodied compliance improves interaction robustness and task adaptability while maintaining manipulation accuracy, providing insights into the role of physical morphology in aerial manipulation.
\end{itemize}

The rest of this work is structured as follows.
Section \uppercase\expandafter{\romannumeral2} reviews the relevant literature.
Section \uppercase\expandafter{\romannumeral3} introduces the system design and implementation, and evaluates the soft structure performance.
Section \uppercase\expandafter{\romannumeral4} presents the system modeling, and Section \uppercase\expandafter{\romannumeral5} describes the composite control framework and the end-effector trajectory generation method.
In Section \uppercase\expandafter{\romannumeral6}, real-world experiments are conducted to explain the capabilities of the proposed system, and Section \uppercase\expandafter{\romannumeral6} provides the discussions of this work.
Finally, Section \uppercase\expandafter{\romannumeral7} concludes this article.

\section{Related Work}

This section reviews representative studies in contact-rich Aerial manipulation, compliance in Aerial Interaction, and variable-stiffness manipulation.
Then, their limitations are discussed to motivate the proposed rigid-soft integrated aerial manipulation framework.

\subsection{Contact-Rich Aerial Manipulation}

Recent advances have evolved aerial manipulation from simple pick-and-place toward increasingly complex physical interaction tasks, including infrastructure inspection, maintenance, assembly, and intervention in challenging environments \cite{2023LiangTMECH,2024LiangTMECH}.
To support these applications, substantial efforts have been devoted to the development of aerial manipulation platforms and control strategies.

From the platform perspective, both underactuated and fully actuated aerial vehicles have been investigated.
Conventional underactuated aerial platforms are attractive due to their mechanical simplicity and high payload efficiency.
However, they suffer from strong coupling between translational and rotational dynamics, which limits their manipulation dexterity.
To overcome these limitations, various fully actuated aerial platforms have been proposed, including tilted \cite{2019RyllIJRR, 2022RashadTRO} configurations and tiltable configurations \cite{2021BodieTRO, 2020AllenspachIJRR}.
By enabling independent control of position and orientation, fully actuated aerial systems significantly improve manipulation dexterity, interaction stability, and workspace accessibility.
In parallel, aerial manipulators have evolved from single-arm configurations \cite{2023LiangTASE,2022LaiTMECH,2019TognonRAL} to dual-arm systems \cite{2017OrsagTRO,2017SuarezIROS} and mechanisms with enlarged workspaces, such as rotary manipulation structures \cite{2024LiangTMECH,2019TrujilloSensors}.

Despite the progress made, contact-rich aerial manipulation still presents fundamentally challenging, as the interaction forces generated at the end-effector are directly coupled to the dynamics of the aerial platform.
While conventional aerial rigid manipulators provide accurate positioning and effective force transmission, they are sensitive to impact disturbances and environmental uncertainties.
Even moderate contact forces can lead to significant attitude deviations, thereby undermining system stability.
To improve interaction performance, existing solutions have explored compliant control strategies, force regulation, or additional force sensing to mitigate interaction disturbances \cite{2019RyllIJRR, 2024LiangTMECH}.
While these approaches can enhance interaction stability under controlled conditions, their effectiveness often require accurate environment modeling and reliable interaction measurements, which may be difficult to obtain in unstructured environments.
Therefore, achieving robust aerial physical interaction and manipulation under uncertain contact conditions remains an open challenge.

\subsection{Compliance in Aerial Interaction}

Compliance has emerged as an effective mechanism for enhancing the safety and robustness of aerial physical interaction.
Compared with purely rigid manipulators, compliant structures can absorb impact energy, accommodate environmental uncertainties, and mitigate the transmission of interaction disturbances to the aerial platform by introducing passive deformation during contact.

Existing studies have explored mechanically compliant aerial manipulators, in which mechanical compliance provides a significant solution by embedding compliant behavior directly into the robot morphology \cite{2024MelletRoboSoft,2015YukselICRA,2026SupaRoboSoft,2016BarteldsRAL, 2018SuarezRAL}.
Several aerial manipulation systems have incorporated flexible joints \cite{2015YukselICRA}, compliant mechanisms \cite{2026SupaRoboSoft,2016BarteldsRAL}, spring-based transmissions \cite{2018SuarezRAL}, and lightweight flexible robotic arms \cite{2024MelletRoboSoft} to improve interaction safety and environmental adaptability. 

Mechanical compliance has been introduced through flexible joints \cite{2015YukselICRA}, compliant mechanisms \cite{2026SupaRoboSoft,2016BarteldsRAL}, spring-based transmissions \cite{2018SuarezRAL}, and lightweight flexible robotic arms \cite{2024MelletRoboSoft}.
These approaches improve interaction safety and passive adaptability during contact.
However, compliance in such systems is typically localized at discrete joints, limiting their ability to realize large and continuous shape adaptation during manipulation.

Recent advances in soft robotics have introduced a new class of aerial manipulators based on soft continuum structures, including pneumatically driven soft robotic arms \cite{2022SzaszRoboSoft}, soft grippers \cite{2021FishmanAC}, and tendon-driven continuum manipulators \cite{2023PengTSMC}.
Unlike rigid manipulators that concentrate compliance at discrete joints, soft robotic systems exhibit intrinsic and distributed compliance along their entire body, enabling safer and more flexible interaction with uncertain or delicate environments while reducing the dependence on precise force regulation and environmental modeling.
This property is particularly valuable for aerial manipulators because it alleviates the burden on sensing, force regulation, and active stabilization.
In this sense, compliance can be viewed as a form of embodied intelligence, where part of the interaction complexity is handled through physical morphology rather than explicit control.

Despite these advantages, purely soft aerial manipulators often suffer from limited load-bearing capability, reduced positioning accuracy, and insufficient structural rigidity during manipulation.
Excessive compliance may lead to large deformations, poor force transmission, and reduced task precision.
As a result, achieving both robust physical interaction and accurate aerial manipulation remains a key challenge for aerial compliant manipulation systems.

\subsection{Variable-Stiffness Manipulation}

The respective limitations of rigid and compliant manipulators have motivated growing interest in variable-stiffness robotic systems.
As an approach to balance the conflicting demands of compliance and precision, variable-stiffness mechanisms have been widely studied in robotics.
In many manipulation tasks, compliance and rigidity are desirable at different stages of interaction.
Low stiffness facilitates contact establishment by enhancing impact absorption, improving interaction safety, and increasing environmental adaptability.
In contrast, high stiffness is often essential for precise positioning, efficient force transmission, and tool operation.

To reconcile these contradictory requirements, various variable-stiffness technologies have been developed.
Common methods involve altering the physical state of materials, such as using low melting point metals \cite{2018HaoJMM} or materials that undergo a glass transition \cite{2008CapadonaScience}.
Chemical methods, like hydrogels \cite{2020ZhuoSciAdv}, have also been employed.
However, these methods typically require heating or electric field actuation, which can complicate integration into drones.
Jamming-based techniques (such as granular jamming \cite{2023AnSoftRobotics, 2016WeiMechatronics, 2013CianchettiIROS}, layer jamming \cite{2019WangTMECH, 2016SantiagoSoftRobotics, 2013KimTRO}, and fiber jamming \cite{2020BrancadoroSR}) provide an effective and convenient way to modify the stiffness of soft robotic arms.
Nonetheless, accommodating the large deformations and irregular shapes of soft arms can make the structural design of such jamming techniques quite complex.
In pursuit of scalable design and manufacturing, chain-mail jamming presents an innovative solution to strengthen soft structures, where the designed three-dimensional structure is wrapped in a flexible membrane and then hardened through vacuum actuation \cite{2021WangNature, 2023XieRAL}.
This solution facilitates the rapid fabrication and reinforcement of soft robotic arms.
The chain-mail structure forms a macroscopic interlocking network when hardened, whose mechanical model more closely resembles an integral beam, which provides significantly higher bending stiffness and load capacity compared to schemes reliant on point-contact force chains (granular jamming) or interlayer friction (layer jamming).
Further, the chain-mail jamming exhibits rapid stiffness switching due to minimal internal material rearrangement and low required vacuum volume.

Variable stiffness has demonstrated considerable success in soft robotics, where stiffness modulation enables robots to transition between highly compliant and rigid configurations.
Such capabilities have been widely exploited in grasping, locomotion, rehabilitation devices, and continuum manipulation.
More recently, researchers have begun exploring the integration of variable-stiffness mechanisms into aerial robotic systems to enhance manipulation performance under diverse operating conditions.
However, the application of variable stiffness in aerial manipulation remains relatively limited.
Aerial robots are subject to stringent constraints on weight, volume, energy consumption, and system complexity, making many existing variable-stiffness technologies difficult to deploy.
Furthermore, the influence of stiffness modulation on the stability of aerial platforms has not yet been systematically investigated.
As a result, developing lightweight variable-stiffness aerial manipulators capable of balancing compliance, precision, and interaction robustness remains an important research challenge.

\begin{figure*}
	\centering
	\includegraphics[width=6.9in]{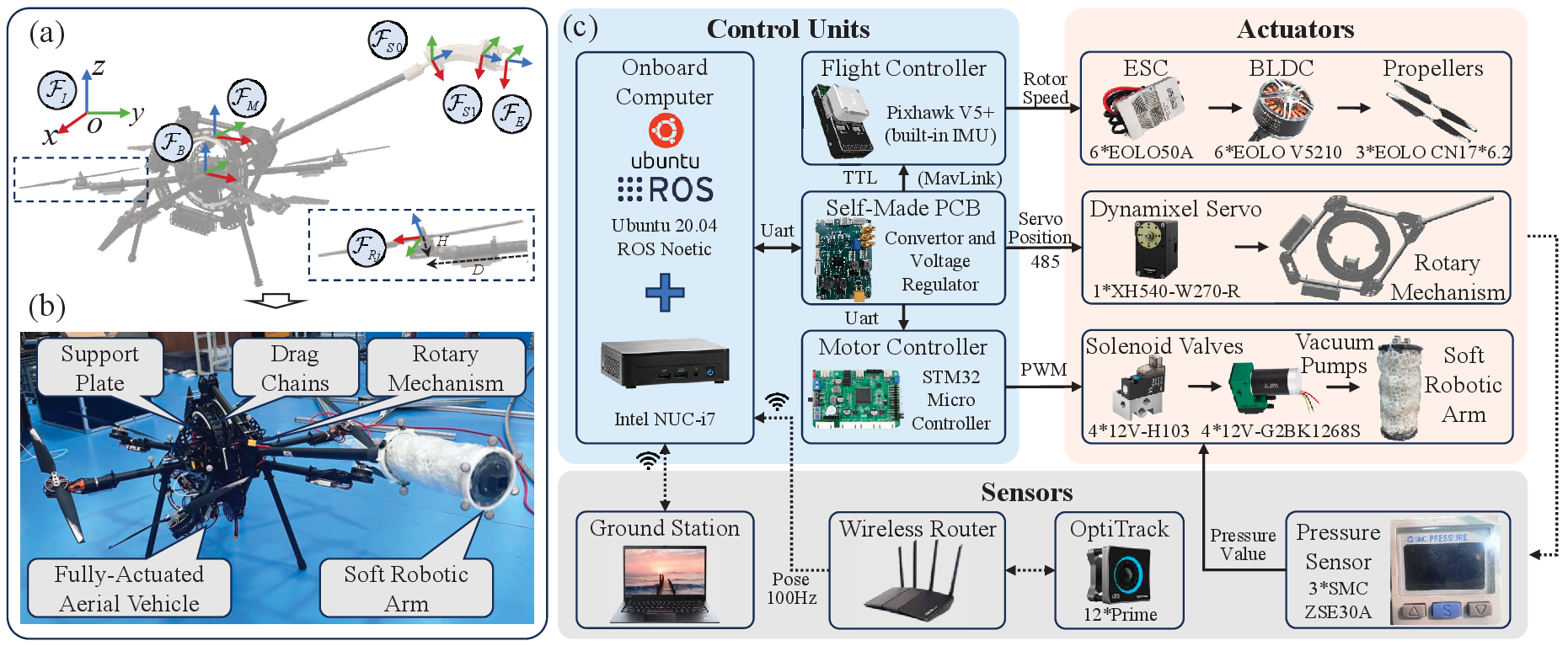}
	\caption{Design and implementation of the AeRSoM robot system.
		(a) Three-dimensional CAD model.
		(b) Real AeRSoM robot consisting of the fully-actuated aerial vehicle, the rotary mechanism, and the soft robotic arm.
		(c) Hardware architecture of the AeRSoM robot system.}
	\label{fig_system}
\end{figure*}

\section{System Overview}

Unlike conventional aerial manipulators that primarily rely on rigid structures and active force regulation, this work investigates embodied compliance as a mechanism for enhancing aerial physical interaction and manipulation.
The proposed AeRSoM robot integrates a fully actuated aerial platform, a rigid-soft manipulator architecture, and a lightweight variable-stiffness mechanism within a unified aerial manipulation framework, aiming to simultaneously achieve interaction robustness, manipulation precision, and environmental adaptability.
The resulting system provides a practical framework for investigating the role of embodied compliance in aerial manipulation and demonstrates how morphology-assisted interaction can enhance the capability of aerial robotic systems operating in uncertain environments.

\subsection{System Design and Implementation}

Following the analysis and discussion of related literature, the AeRSoM robot system is designed and implemented in this part, as illustrated in Fig. \ref{fig_system}.
Initially, the three-dimensional (3D) CAD model of the robot is designed and shown in Fig. \ref{fig_system}(a), which consists of a fully-actuated flight platform, a one-DOF robotic rotary mechanism, and a segment of a lightweight soft robotic arm.
According to the designed CAD model, the real AeRSoM robot is assembled in Fig. \ref{fig_system}(b).
The rotary mechanism can achieve circular motion in the $x$-$z$ plane with respect to the floating base through the gear transmission mechanism.
The main body of the rotary mechanism is a large transmission gear with a 360$^\circ$ gear chute embedded inside, and the chute is connected to a small gear on the Dynamixel XH540-W270-R servo, where the transmission ratio is approximately 1:5.
To avoid the chattering of the rotary mechanism in the lateral direction, two support plates are designed to connect the base of the rotary mechanism and the flight vehicle base.
Additionally, the drag chains are installed on the rotary mechanism to arrange the wiring of the electronic devices on the soft robotic arm and the end-effector.
The hardware architecture is depicted in Fig. \ref{fig_system}(c), where the total weight is approximately 8.875 kg.
The fully-actuated aerial vehicle is developed from a traditional hexarotor with the symmetrical rotor spacing of 1200 mm, where propellers are tilted with a fixed angle $\alpha$ ($\alpha = 30^\circ$) and point in different directions.
An Intel NUC-i7 onboard computer is integrated into the flight base to process high-level signals, while a Pixhawk V5+ is responsible for managing the low-level flight control.
The commanded rotor speed signal generated by the flight controller is sent to the EOLO50A electronic speed controller (ESC), which drives the EOLO V5210 BLDC motors equipped with EOLO CN17$\ast$6.2 inch propellers to produce flight motion.
The robotic rotary mechanism is powered by a Dynamixel XH540-W270-R servo through a gear transmission system.
To maintain a lightweight design without compromising strength, the rotary mechanism is constructed using carbon fiber materials and aluminum alloys, while the transmission components are made of 3D-printed plastic materials.

\begin{figure}
	\centering
	\includegraphics[width=3.3in]{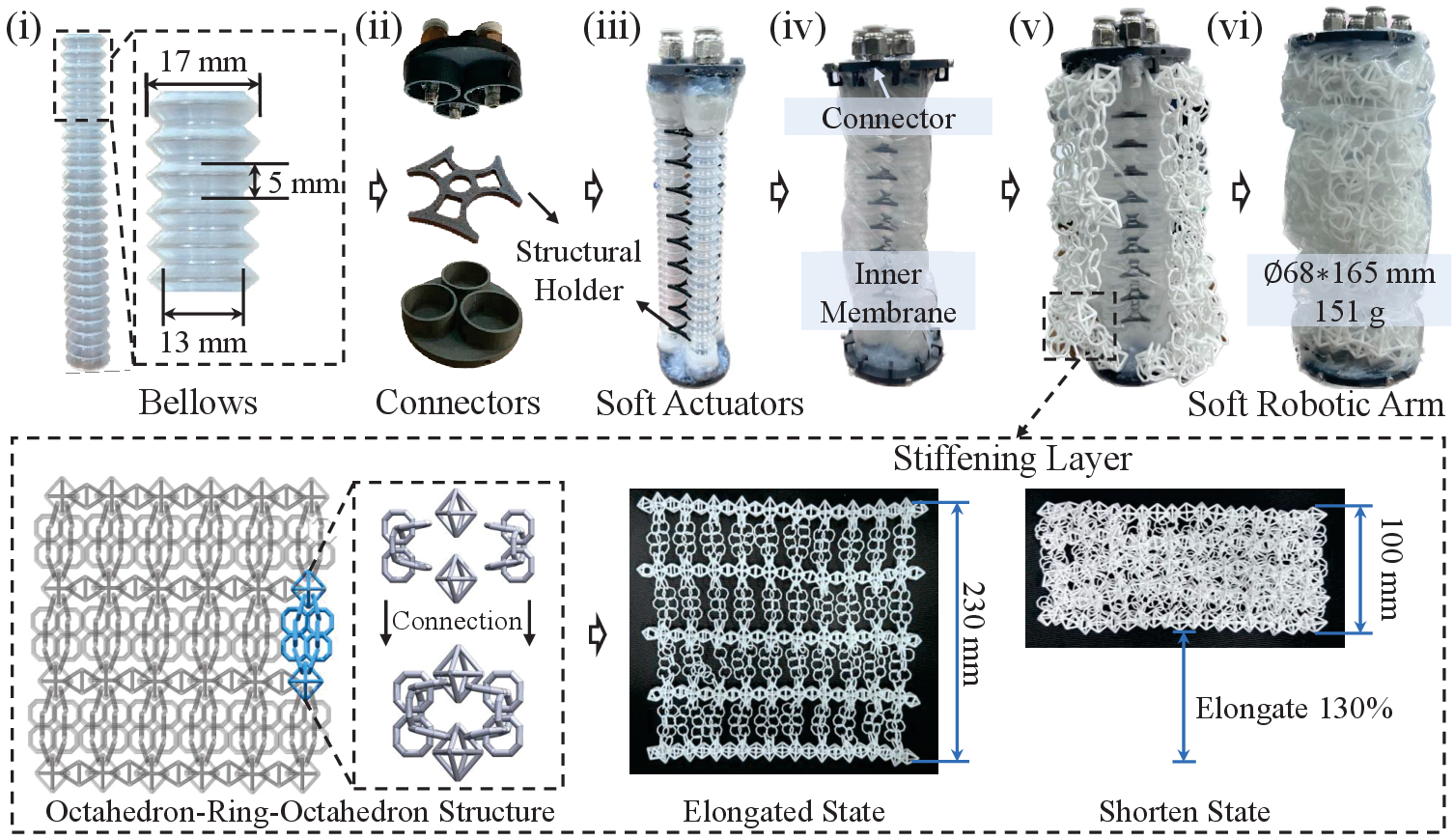}
	\caption{Fabrication process and assembly of the soft robotic arm.
		(i) Selection of the bellows.
		(ii) Design of the connectors and structural holders.
		(iii) Assembly of three soft actuators.
		(iv) Containing the soft actuators with a soft inner membrane.
		(v) Fabrication and assembly of the stiffening layer.
		(vi) Assembly of the soft robotic arm, where the stiffening layer is sealed with soft membranes.}
	\label{fig_softArm}
\end{figure}

\begin{figure}
	\centering
	\includegraphics[width=3.3in]{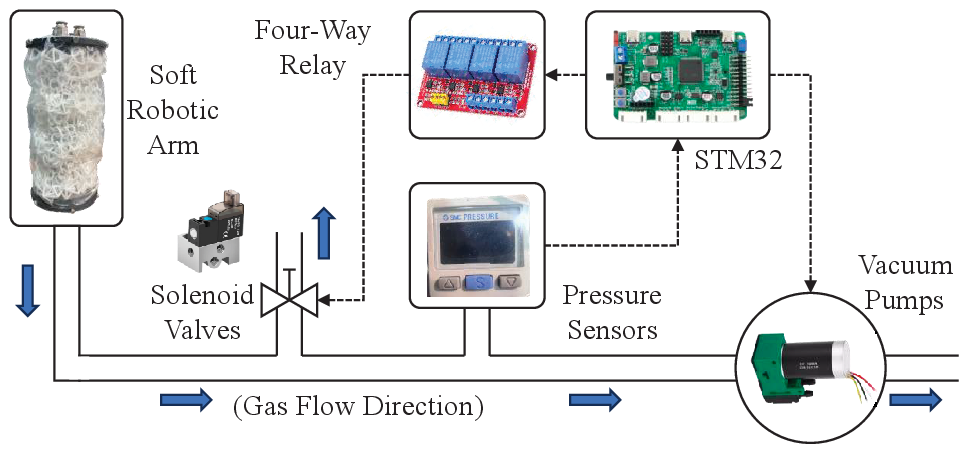}
	\caption{Schematic of the pneumatic control system for the soft robotic arm.}
	\label{fig_pneumaticControlSystem}
\end{figure}

The lightweight soft robotic arm is fabricated with three soft actuators along with a stiffening layer, as shown in Fig. \ref{fig_softArm}.
First, three polyethylene bellows serve as pneumatic actuators for the soft robotic arm, and each bellows is depressurized to shorten its length to generate motion (Fig. \ref{fig_softArm}(i)).
Then, the connectors and the structural holders are 3D-designed and printed, where the structural holders can ensure that the axes of the parallel actuator units can remain relatively parallel during the shortening process (Fig. \ref{fig_softArm}(ii)).
Subsequently, three soft actuators are arranged in a circle and assembled with connectors and structural holders (Fig. \ref{fig_softArm}(iii)), and then the soft actuators are assembled with a soft inner membrane (Fig. \ref{fig_softArm}(iv)).
Next, employing the design principles of chain-mail jamming \cite{2021WangNature, 2023XieRAL}, we create a 3D CAD model of the stiffening layer that features the octahedron-ring-octahedron connected structure, and the elongation of this structure is proved to be 130$\%$.
Then, the 3D-printed stiffening layer is assembled outside the inner soft membrane (Fig. \ref{fig_softArm}(v)).
Ultimately, a lightweight soft robotic arm is fabricated and assembled by combining the sealed stiffening layer and soft actuators (Fig. \ref{fig_softArm}(vi)), where the sealed stiffening layer is simply operated using a vacuum.
Without actuation, the size of the soft robotic arm without end-effectors is ${\emptyset}$68$\ast$165 mm, and the total weight of the soft robotic arm is about 151 g.
From Fig. \ref{fig_system}(c), the vacuum pressures produced by 12V-G2BK1268S vacuum pumps act on soft actuators to enable motion in space.
To implement the control of the pneumatic pressures in the chambers, 12V-H103 micro ON-OFF solenoid valves and a four-way relay are employed to regulate the pressure within the chamber, and the pneumatic pressures in the actuators are monitored using SMC ZSE30A pressure sensors.
A STM32 microcontroller is utilized for the low-level process, while high-level control commands are issued from the Intel onboard computer.
The schematic of the pneumatic control system for the soft robotic arm is shown in Fig. \ref{fig_pneumaticControlSystem}.

It is worth noting that a printed circuit board (PCB) is designed and manufactured to manage power supply and facilitate signal conversion.
The control units in Fig. \ref{fig_system}(c) detail the signal conversion process in conjunction with other controllers or servos, which the PCB is responsible for converting and supplying the appropriate voltage to support both the onboard computer and servos.

\subsection{Soft Structure Performance Evaluation}

\begin{figure}
	\centering
	\includegraphics[width=3.4in]{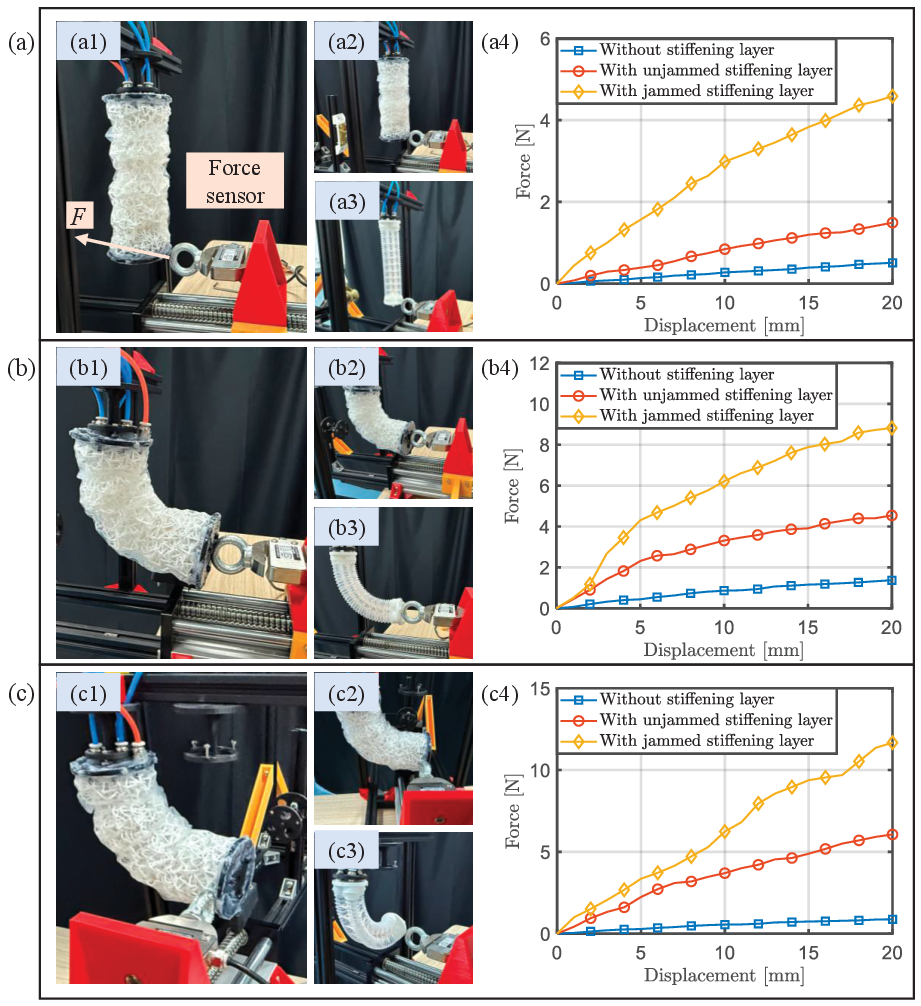}
	\caption{Stiffness evaluation of the soft robotic arm at three different states: (a) free-hanging, (b) axial when 90$^\circ$ bending, and (c) tangential when 90$^\circ$ bending.
		Snapshots of the soft robotic arm under three different cases: with the jammed stiffening layer ((a1), (b1), and (c1)), with the unjammed stiffening layer ((a2), (b2), and (c2)), without the stiffening layer ((a3), (b3), and (c3)).
		Force-displacement curves at (a4) the free-hanging state, (b4) the axial state when 90$^\circ$ bending, and (c4) the tangential state when 90$^\circ$ bending.}
	\label{fig_stiffnessTest}
\end{figure}

This subsection will evaluate the stiffness and load capacity of the fabricated soft robotic arm, as tools need to be mounted on the end-effector to perform aerial manipulation tasks.
To ensure consistency and fairness in the upcoming capacity evaluations, the same pneumatic pressures are applied to the soft actuators during each test.
Consequently, the kinematic modeling and control of the soft arm can be set aside for now, but these topics will be addressed in the following section.

\begin{table}
	\centering
	\renewcommand{\arraystretch}{1.2}
	\caption{Comparison of different soft robotic arms with variable stiffness capability
		\label{Table_ComparisonJamming}}
	\begin{tabular}{c|c|c|c}
		\toprule
		\textbf{Arms}          & \textbf{Stiffening Principle} & \textbf{\begin{tabular}[c]{@{}c@{}}Stiffening Work\\ Range\end{tabular}} & \textbf{\begin{tabular}[c]{@{}c@{}}Maximum\\ Stiffening Rate\end{tabular}} \\ \midrule
		\textbf{\begin{tabular}[c]{@{}c@{}}This\\ work\end{tabular}}     & \textbf{Chain-mail}   & \textbf{All directions}            & \textbf{1230\%}                  \\ \hline
		\cite{2023AnSoftRobotics}         & Granular jamming              & All directions                     & 750\%                            \\ \hline
		\cite{2016WeiMechatronics}        & Granular jamming              & All directions                     & 900\%                            \\ \hline
		\cite{2013CianchettiIROS} & Granular jamming              & \begin{tabular}[c]{@{}c@{}}Mainly increase\\ in-plane stiffness\end{tabular} & 36\%                             \\ \hline
		\cite{2019WangTMECH}       & Layer jamming                 & All directions                     & 700\%                            \\ \hline
		\cite{2016SantiagoSoftRobotics}   & Layer jamming                 & \begin{tabular}[c]{@{}c@{}}Mainly increase\\ in-plane stiffness\end{tabular} & 46\%                             \\ \hline
		\cite{2013KimTRO}        & Layer jamming                 & All directions                     & 90\%                             \\ \bottomrule
	\end{tabular}
\end{table}

The stiffness of the soft robotic arm is tested under three different cases: the soft robotic arm without the stiffening layer, with the unjammed stiffening layer, and with the jammed stiffening layer, as shown in Fig. \ref{fig_stiffnessTest}.
For each case, the stiffness tests are performed at three different states: free-hanging (Fig. \ref{fig_stiffnessTest}(a)), axial when 90$^\circ$ bending (Fig. \ref{fig_stiffnessTest}(b)), and tangential when 90$^\circ$ bending (Fig. \ref{fig_stiffnessTest}(c)), in which a force sensor is placed on a precisely controlled lead screw slide to push the soft robotic arm in 1 mm increments, and then the applied forces are recorded and plotted with the displacement (Fig. \ref{fig_stiffnessTest}(a4), Fig. \ref{fig_stiffnessTest}(b4), and Fig. \ref{fig_stiffnessTest}(c4)).
As shown in Fig. \ref{fig_stiffnessTest}(a), the stiffness of the soft robotic arm without the stiffening layer in the free-hanging state is extremely low, and the applied force is only 0.511 N when pushing a distance of 20 mm.
After integrating the unjammed stiffening layer, the stiffness is not significantly improved, and the applied force at a pushing distance of 20 mm is 1.491 N.
When the stiffening layer is jammed, the applied force comes to 4.583 N, which is nearly 9 times that of the soft robotic arm without the stiffening layer.
Then, for the 90$^\circ$ bending state, the axial stiffness of the soft robotic arm without the stiffening layer is slightly increased due to the introduction of pneumatic pressure, applying a force of 1.373 N at a displacement of 20 mm, as exhibited in Fig. \ref{fig_stiffnessTest}(b).
The unjammed and jammed stiffening layers generate 4.542 N and 8.817 N, respectively, almost 6.5 times that of the original soft robotic arm.
Finally, since the soft materials generally cannot withstand much out-of-plane force, the tangential stiffness at the 90$^\circ$ bending state is tested.
As plotted in Fig. \ref{fig_stiffnessTest}(c), the applied force of the soft robotic arm without the stiffening layer is 0.879 N at a displacement of 20 mm.
The unjammed stiffening layer generates 6.063 N, and the jammed stiffening layer significantly copes with this drawback by increasing the force to 11.671 N, almost 13.3 times larger.
Table \ref{Table_ComparisonJamming} concludes a quantitative comparison with other jamming methods (such as granular jamming and layer jamming).

\begin{figure}
	\centering
	\includegraphics[width=3.3in]{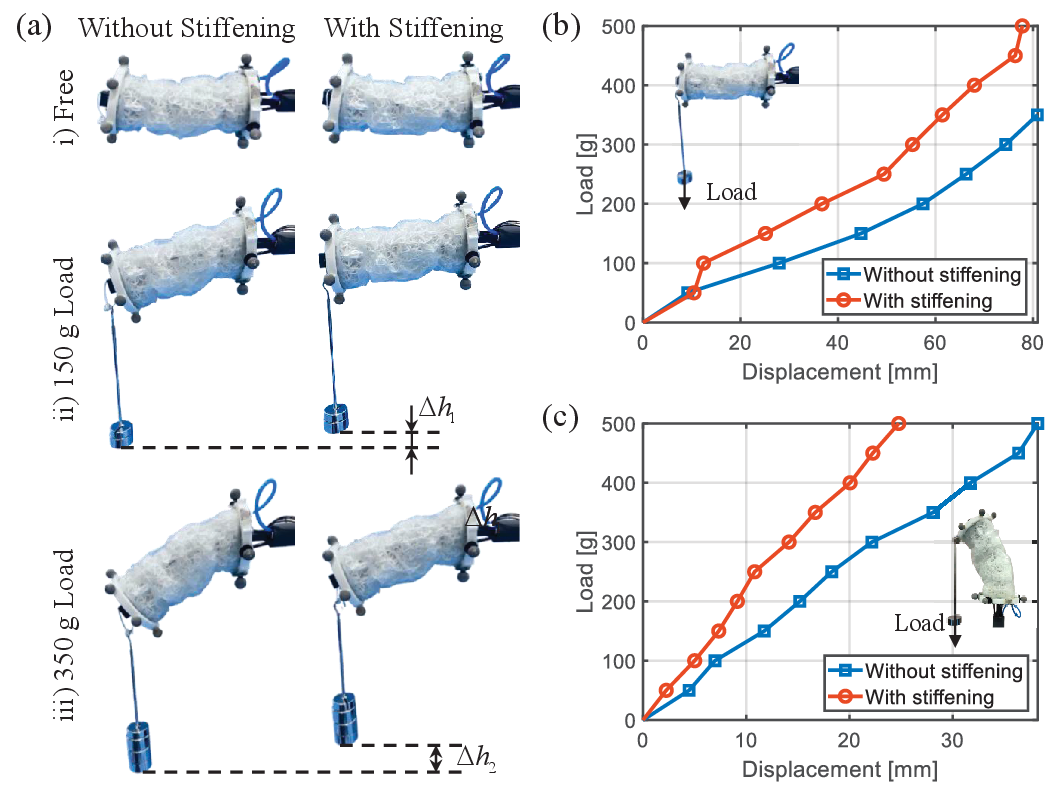}
	\caption{Load capacity evaluation of the soft robotic arm.
		(a) The soft robotic arm is placed horizontally under the same actuation with different loads ((i) free load, (ii) 150 g, and (iii) 350 g), and the deviation of the soft robotic arm in the $z$-direction in the inertial frame $\mathcal{F}_I$ slightly decreases after stiffening compared with the unjammed state ($\Delta h_1 = 16$ mm and $\Delta h_2 = 20$ mm).
		(b) and (c) The relationship between the load weight and displacement in the $z$-direction in $\mathcal{F}_I$ with and without stiffening, where the soft arm is place horizontally (b) and vertically (c).}
	\label{fig_softPerformance}
\end{figure}

The load capacity tests of the soft robotic arm are conducted, and the evaluated results are depicted in Fig. \ref{fig_softPerformance}.
To account for the gravitational effects on the soft robotic arm, it is initially positioned as a horizontal cantilever beam (Fig. \ref{fig_softPerformance}(a)-(i)).
Compared with the unjamming state, the deviation of the soft robotic arm in the $z$-direction in the inertial frame $\mathcal{F}_I$ slightly decreases after stiffening under the same actuation and load conditions.
When a load of 150 g is applied, the stiffened soft arm droops, resulting in a height difference of $\Delta h_1$ (16 mm) in comparison to its unjamming state (Fig. \ref{fig_softPerformance}(a)-(ii)).
As the load increases to 350 g, the height difference grows slightly to $\Delta h_2$ (20 mm) (Fig. \ref{fig_softPerformance}(a)-(iii)).
If the load exceeds 350 g, applying additional weight to the unjammed soft arm could cause damage, but the stiffened state remains within its load capacity.
Fig. \ref{fig_softPerformance}(b) illustrates the relationship between the load weight and displacement of the soft arm, both in its jammed and unjammed states, with load values ranging from free load up to 500 g.
Fig. \ref{fig_softPerformance}(c) is similar to Fig. \ref{fig_softPerformance}(b), recording the relationship between the load weight and displacement of the soft arm in the jammed and unjammed states when placed vertically.
These tests demonstrate that the stiffened soft robotic arm can support heavier loads without large deviations, making it beneficial for practical applications.

\section{Modeling}

\begin{table}
	\centering
	\renewcommand{\arraystretch}{1.2}
	\begin{threeparttable}
		\caption{Nomenclature}
		\label{Table_Symbol}
		\centering
		\begin{tabular}{l l}
			\toprule
			\textbf{Symbols} & \textbf{Definitions} \\
			\midrule
			Frames & \\
			$\mathcal{F}_I$ & Inertial frame \\
			$\mathcal{F}_B$ & Body frame of the aerial vehicle \\
			$\mathcal{F}_M$ & Joint frame of the rotary mechanism \\
			$\mathcal{F}_{S0}$ & Base frame of the soft robotic arm \\
			$\mathcal{F}_{S1}$ & End frame of the soft robotic arm \\
			$\mathcal{F}_E$ & End-effector frame \\
			Manipulator & \\
			$u_1$, $u_2$, $u_3 \in \mathbb{R}$ & $i$-th chamber pressure ($i = 1, 2, 3$) \\
			$u_s \in \mathbb{R}$ & Stiffening layer pressure \\
			$l_1$, $l_2$, $l_3 \in \mathbb{R}$ & $i$-th chamber length ($i = 1, 2, 3$) \\
			$L \in \mathbb{R}$ & Length of the soft robotic arm \\
			$\phi$, $\theta \in \mathbb{R}$  & Curvature angle and direction of bending \\
			$\rho \in \mathbb{R}$ & Curvature of the soft robotic arm \\
			$r \in \mathbb{R}$ & Curvature radius of the soft robotic arm \\
			$h \in \mathbb{R}$ & Cross-sectional radius \\
			Aerial vehicle &   \\
			$\alpha \in \mathbb{R}$ & Tilted angle of each propeller ($\alpha = 30^\circ$) \\
			$m_s \in \mathbb{R}$ & Total mass of the whole system \\
			$\bm J_b \in \mathbb{R}^{3 \times 3}$ & Inertial matrix of the vehicle \\
			$\bm p_b^I, \bm v_b^I \in \mathbb{R}^3$ & Vehicle's position and linear velocity in $\mathcal{F}_I$ \\
			$\bm v_b, \bm \omega_b \in \mathbb{R}^3$ & Twist of the vehicle in $\mathcal{F}_B$ \\
			$\bm R_\ast^\star \in \mathbb{R}^{3 \times 3}$ & Rotation matrix from $\mathcal{F}_\ast$ to $\mathcal{F}_\star$ \\
			$g \in \mathbb{R}$ & Gravity constant ($g = 9.81$ kg$\cdot$m/s$^2$) \\
			$\bm e_3 \in \mathbb{R}^3$ & Unit vector ($\bm e_3 = [0, 0, 1]^\top$) \\
			\bottomrule
		\end{tabular}
	\end{threeparttable}
\end{table}

The model of the AeRSoM robot is shown in Fig. \ref{fig_system}(a), six coordinate frames are defined to describe the kinematics of the AeRSoM system:
the inertial frame $\mathcal{F}_I$,
the body frame of the aerial vehicle $\mathcal{F}_B$,
the joint frame of the rotary mechanism $\mathcal{F}_M$,
the base frame $\mathcal{F}_{S0}$ and end frame $\mathcal{F}_{S1}$ of the soft robotic arm,
and the end-effector frame $\mathcal{F}_E$.
The symbols are summarized in Table \ref{Table_Symbol} for the convenience of reading.

\subsection{Forward Kinematic Model}

Based on the defined coordinate frames of the robot system, the forward kinematic model is derived to calculate the spatial position and orientation of the end-effector, taking into account the motion of both the flight platform and the manipulator.
The forward kinematics of the AeRSoM robot is described as

\begin{align}
	\label{Eq.ForwardKin.1-1}
	\bm T_E^I & = \bm T_B^I \bm T_M^B \bm T_E^M
	= \left[\begin{array}{cc}
		\bm R_E^I & \bm p_e^I \\
		\bm 0 & 1
	\end{array}\right] \\
	\label{Eq.ForwardKin.1-2}
	\bm R_E^I & = \bm R_B^I \bm R_M^B \bm R_E^M, ~~ \bm p_e^I = \bm p_b^I + \bm R_B^I (\bm p_m^b + \bm R_M^B \bm p_e^m)
\end{align}
where $\bm T_\ast^\star \in \mathbb{R}^{4 \times 4}$ and $\bm R_\ast^\star \in \mathbb{R}^{3 \times 3}$ represent the transformation and rotation matrices from the frame $\ast$ to the frame $\star$, respectively, where $\ast = \{B, M, E\}$ and $\star = \{I, B, M\}$.
The vectors $\bm p_e^I = [x_e, y_e, z_e]^\top \in \mathbb{R}^3$ and $\bm p_b^I \in \mathbb{R}^3$ denote the end-effector position and vehicle position expressed in the inertial frame $\mathcal{F}_I$, respectively.
The vectors $\bm p_m^b \in \mathbb{R}^3$ and $\bm p_e^m \in \mathbb{R}^3$ represent the position of the rotary mechanism expressed in the body frame $\mathcal{F}_B$ and the end-effector position in $\mathcal{F}_M$, respectively.

\begin{figure}
	\centering
	\includegraphics[width=3.3in]{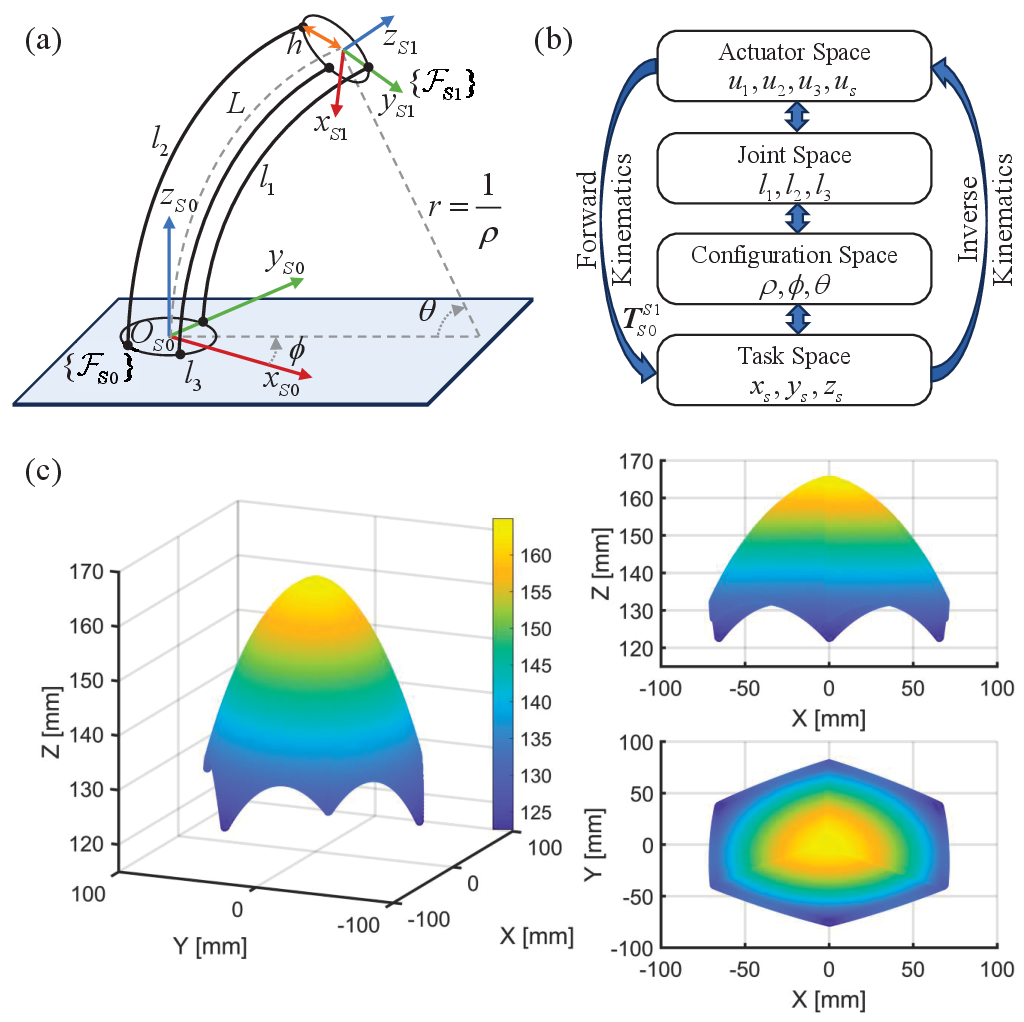}
	\caption{(a) Frame transformation between the base and end of the soft robotic arm.
		(b) The mapping among the actuator space, joint space, configuration space, and task space.
		(c) Simulated workspace results of the soft robotic arm.}
	\label{fig_forwardKin}
\end{figure}

Note that the transformation matrix $\bm T_E^M = \bm T_{S0}^M \bm T_{S1}^{S0} \bm T_E^{S1}$ contains the transformation between the base and end of the soft robotic arm $\bm T_{S1}^{S0}$, which needs to be determined.
The matrix $\bm T_{S0}^M$ is the transformation matrix between the joint frame $\mathcal{F}_M$ and the base frame of the soft robotic arm $\mathcal{F}_{S0}$, and $\bm T_E^{S1}$ is the transformation matrix between the end-effector frame $\mathcal{F}_E$ and the end frame of the soft robotic arm $\mathcal{F}_{S1}$.
Fig. \ref{fig_forwardKin}(a) illustrates the modeling of the soft robotic arm, where three chambers of the soft robotic arm are assembled to be parallel, and thus the soft robotic arm is assumed to have constant curvature.
To acquire the forward kinematics of the soft robotic arm, the mapping from the actuator space $\{u_1, u_2, u_3, u_s\}$ to the task space $\{x_s, y_s, z_s\}$ of the soft robotic arm needs to be determined, as illustrated in Fig. \ref{fig_forwardKin}(b), where the stiffening layer pressure $u_s$ is isolated from the chamber pressures $\{u_1, u_2, u_3\}$ and only used to enable the stiffening layer.
From Fig. \ref{fig_forwardKin}(a), the transformation from the joint space $\{l_1, l_2, l_3\}$ to configuration space $\{\rho, \phi, \theta\}$ is expressed as
\begin{align}
	\label{Eq.ForwardKin.2}
	\rho(l_i) & = \frac{1}{r} = \frac{2 \sqrt{l_1^2 + l_2^2 + l_3^2 - l_1 l_2 - l_1 l_3 - l_2 l_3}}{h (l_1 + l_2 + l_3)} \\
	\label{Eq.ForwardKin.3}
	\phi(l_i) & = \tan^{-1} \left( \frac{\sqrt{3} (l_2 + l_3 - 2 l_1)}{3 (l_2 - l_3)} \right) \\
	\label{Eq.ForwardKin.4}
	\theta(l_i) & = \frac{2 \sqrt{l_1^2 + l_2^2 + l_3^2 - l_1 l_2 - l_1 l_3 - l_2 l_3}}{3 h}
\end{align}
where $l_i$ represents the $i$-th chamber length;
$\rho$, $\phi$, and $\theta$ are the curvature, curvature angle, and bending angle of the soft robotic arm, respectively;
$h$ denotes the cross-sectional radius.
For modeling the transformation $\bm T_{S0}^{S1}$ from the configuration space $\{\rho, \phi, \theta\}$ to the task space $\{x_s, y_s, z_s\}$, firstly, the arm rotates with the angle $\theta$ around $y$-axis of $\mathcal{F}_{S0}$, i.e., $\bm R_y(\theta)$;
subsequently, the arm rotates with the angle $\phi$ around $z$-axis of $\mathcal{F}_{S0}$, i.e., $\bm R_z(\phi)$, and then the arm is moved out of the $x$-$z$ plane with the translation $\bm p_\theta = r [1 - c_\theta, 0, s_\theta]^\top$;
finally, the posture is adjusted by right-multiplying the rotation matrix $\bm R_z(-\phi)$.
The transformation matrix $\bm T_{S0}^{S1}$ is expressed as
\begin{equation}
	\begin{aligned}
		\label{Eq.ForwardKin.5}
		\bm T_{S0}^{S1} & = \left[\begin{array}{cc}
			\bm R_z(\phi) & \bm 0 \\
			\bm 0 & 1
		\end{array}\right]
		\left[\begin{array}{cc}
			\bm R_y(\theta) & \bm p_\theta \\
			\bm 0 & 1
		\end{array}\right]
		\left[\begin{array}{cc}
			\bm R_z(-\phi) & \bm 0 \\
			\bm 0 & 1
		\end{array}\right] \\
		& = \left[\begin{array}{cc}
			\bm R_{S0}^{S1} & \bm p_{S0}^{S1} \\
			\bm 0 & 1
		\end{array}\right]
	\end{aligned}
\end{equation}
\begin{equation}
	\begin{aligned}
		\label{Eq.ForwardKin.6}
		\bm R_{S0}^{S1} =
		\left[
		\begin{array}{ccc}
			c_\phi^2 (c_\theta - 1) + 1 & s_\phi c_\phi (c_\theta - 1) & c_\phi s_\theta \\
			s_\phi c_\phi (c_\theta - 1) & s_\phi^2 (c_\theta - 1) + 1 & s_\phi s_\theta \\
			-c_\phi s_\theta & -s_\phi s_\theta & c_\theta \\
		\end{array}
		\right]
	\end{aligned}
\end{equation}
\begin{equation}
	\begin{aligned}
		\label{Eq.ForwardKin.7}
		\bm p_{S0}^{S1} = [x_s, y_s, z_s]^\top \triangleq r \left[c_\phi (1 - c_\theta), s_\phi (1 - c_\theta), s_\theta \right]^\top
	\end{aligned}
\end{equation}
where $c_\phi$, $s_\phi$, $c_\theta$, $s_\theta$ represent $\cos \phi$, $\sin \phi$, $\cos \theta$, $\sin \theta$, respectively.
$\bm p_{S0}^{S1}$ and $\bm R_{S0}^{S1}$ denote the translational motion and rotation of the soft robotic arm expressed in $\mathcal{F}_{S0}$, respectively.

\subsection{Inverse Kinematic Model}

The inverse kinematics of the AeRSoM robot can be used to calculate the movement of the actuator parts associated with the desired pose of the tool's end-effector by using \eqref{Eq.ForwardKin.1-2}.
In particular, the rigid portion of the AeRSoM robot can be easily resolved according to repeated measurements and tests.
To determine the end motion of the soft robotic arm, the inverse kinematics model is established based on the given end position $\{x_s, y_s, z_s\}$.
The modeling procedures are given in Fig. \ref{fig_forwardKin}(b): the first is to transform from the given end position $\{x_s, y_s, z_s\}$ to the arc parameter $\{\rho, \phi, \theta\}$ according to the geometric calculations;
the second is to transform from the arc parameter $\{\rho, \phi, \theta\}$ to the chamber length $\{l_1, l_2, l_3\}$.
According to geometric calculations using \eqref{Eq.ForwardKin.7}, the arc parameter $\{\rho, \phi, \theta\}$ can be obtained by the given end position as
\begin{align}
	\label{Eq.IKArm.1}
	\phi & = \tan^{-1} \left( \frac{y_s}{x_s} \right) \\
	\label{Eq.IKArm.2}
	\rho & = \frac{1}{r} = \frac{2 \sqrt{x_s^2 + y_s^2}}{x_s^2 + y_s^2 + z_s^2} \\
	\label{Eq.IKArm.3}
	\theta & = \cos^{-1} \left( 1 - \rho \sqrt{x_s^2 + y_s^2} \right)
\end{align}

Notice that three chambers actuated simultaneously could result in the singular situation ($l_1 = l_2 = l_3$), which indicates the only contraction motion of the soft robotic arm.
Further, considering the general bending case, the chamber length $\{l_1, l_2, l_3\}$ can be obtained by geometric calculation \cite{2021GongIJRR}.
The installed location of three chambers are shown in Fig. \ref{fig_forwardKin}(a).
The direction from the origin of the coordinate system to the first actuator is the positive direction of the $y_{S0}$-axis, and the second and third actuators are placed in a circular order with equal distances.
Based on the geometric relationship in Fig. \ref{fig_forwardKin}(a), the length of each chamber can be computed and represented with respect to arc parameters $\{\rho, \phi, \theta\}$ as
\begin{equation}
	\begin{aligned}
		\label{Eq.IKArm.4}
		l_i = r \theta - \theta h \cos \left[ \frac{2 \pi}{3} (i - 1) + \frac{\pi}{2} - \phi \right], ~ i = 1, 2, 3
	\end{aligned}
\end{equation}

Therefore, the inverse transformation from the given end position $\{x_s, y_s, z_s\}$ to the chamber length $\{l_1, l_2, l_3\}$ can be obtained.
Then, the workspace of the soft robotic arm can be evaluated according to the kinematic model, as depicted in Fig. \ref{fig_forwardKin}(c), where the workspace is 138 mm in length, 158 mm in width, and 42 mm in height.

\subsection{Dynamic Model}

In the previous work involving rotary mechanism-based aerial manipulators, the manipulator is assumed to be in quasi-static motion, and the dynamics of the manipulator is treated as a small disturbance to the flight platform compensated by the platform controller.
Considering dynamic motion, the strong coupling disturbances between the manipulator and the flight platform cannot be negligible.
Therefore, this work incorporates a comprehensive dynamic model of the entire system.
Except for the soft robotic arm, the aerial body and other components are assumed to be rigid.
Using the Newton-Euler equations, the flight dynamics can be modeled in the body frame $\mathcal{F}_B$ as
\begin{equation}
	\begin{aligned}
		\label{Eq.FlightDyn.1}
		\bm M \dot{\bm t}_b + \bm C \bm t_b + \bm G = \bm W_c + \bm W_m + \bm W_e
	\end{aligned}
\end{equation}
where $\bm M \in \mathbb{R}^{6 \times 6}$ represents the positive definite inertia matrix, $\bm C \in \mathbb{R}^{6 \times 6}$ involves the centrifugal and Coriolis terms, and $\bm G \in \mathbb{R}^6$ is the gravity term.
The vector $\bm t_b = [\bm v_b^\top, \bm \omega_b^\top]  \in \mathbb{R}^6$ denotes the twist of the aerial vehicle, where $\bm v_b \in \mathbb{R}^3$ and $\bm \omega_b \in \mathbb{R}^3$ are the linear and angular velocities of the vehicle expressed in the body frame $\mathcal{F}_B$, respectively.
The terms $\bm W_c \in \mathbb{R}^6$ and $\bm W_e \in \mathbb{R}^6$ denote the control and external wrenches acting on the flight platform, respectively.
The term $\bm W_m \in \mathbb{R}^6$ is the coupling disturbance from the manipulator acting on the flight platform.
More specifically, the matrices expressed in the body frame $\mathcal{F}_B$ are described as
\begin{equation}
	\begin{aligned}
		\label{Eq.FlightDyn.2}
		\bm M & = \mathrm{diag} \left(
		\left[
		\begin{array}{cc}
			m_s \bm I_{3 \times 3} & \bm J_b \\
		\end{array}
		\right]
		\right) \\
		\bm C & = \mathrm{diag} \left(
		\left[
		\begin{array}{cc}
			m_s [\bm \omega_b]_\times & -[\bm J_b \bm \omega_b]_\times \\
		\end{array}
		\right]
		\right) \\
		\bm G & = \left[
		\begin{array}{cc}
			m_s (\bm R_I^B g \bm e_3)^\top & \bm 0 \\
		\end{array}
		\right]^\top
	\end{aligned}
\end{equation}
where $m_s$ and $\bm J_b \in \mathbb{R}^{3 \times 3}$ are the total mass of the system and the inertial matrix of the aerial vehicle, respectively;
$g$ is the gravity constant, and $\bm e_3 = [0,0,1]^\top$;
the symbol $[\cdot]_\times$ denotes the skew-symmetric matrix;
$\bm R_I^B \in SO(3)$ is the rotation matrix from the inertial frame $\mathcal{F}_I$ to the body frame $\mathcal{F}_B$.

Rotor groups generate flight motion by receiving speed signals from the flight controller.
Therefore, the control wrench $\bm W_c$ of the system is converted into the thrust and torque generated by each rotor, which is associated with the speed of each rotor.
As shown in Fig. \ref{fig_system}(a), define $\mathcal{F}_{Rj}: \{O_{Rj} - x_{Rj} y_{Rj} z_{Rj}\}$ related to the the $j$-th rotor, and $z_{Rj}$ coincides with the thrust direction.
Then, the position and orientation of the $j$-th rotor-propeller group relative to the body frame of the aerial vehicle $\mathcal{F}_B$ are expressed as
\begin{align}
	\label{Eq.FlightDyn.3}
	\bm p_{Rj}^B & = \bm R_z \left( (j - 1) \frac{\pi}{3} \right) \bm D + \bm R_x \left( (-1)^{j+1} \alpha \right) \bm H, \\
	\label{Eq.FlightDyn.4}
	\bm R_{Rj}^B & = \bm R_z \left( (j - 1) \frac{\pi}{3} \right) \bm R_x \left( (-1)^{j+1} \alpha \right), ~ j = 1, \cdots, 6
\end{align}
where $\bm R_x(\cdot)$ and $\bm R_z(\cdot) \in SO(3)$ denote the rotation matrices about the $x$ and $z$ axes, respectively;
$\bm D$, $\bm H \in \mathbb{R}^3$ are the vectors from the center of the aerial vehicle and the rotor-propeller group to the center of the tilting rotation, respectively.

\begin{figure*}
	\centering
	\includegraphics[width=6.0in]{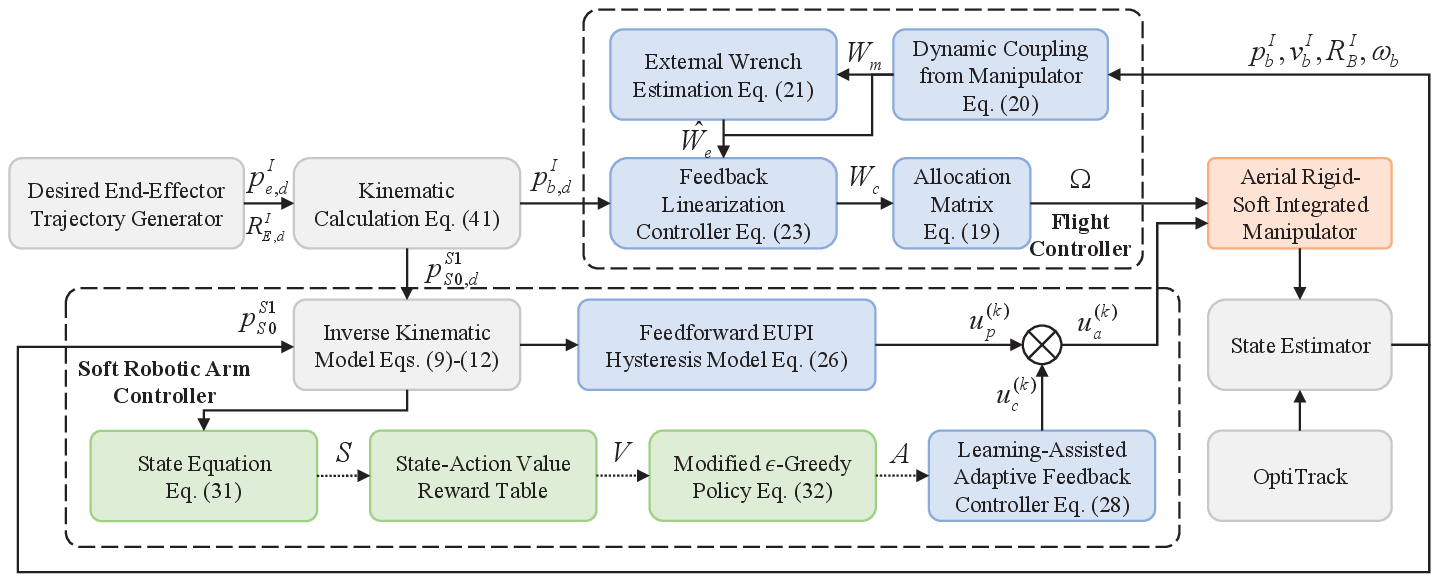}
	\caption{Block diagram of the proposed composite control framework of the AeRSoM robot containing the robust flight controller and the RL-Based adaptive soft robotic arm controller.}
	\label{fig_controller}
\end{figure*}

The rotating propeller generates thrust and torque at the central point $O_{Rj}$.
By approximation, both thrust and torque can be modeled by the square of the rotational speed as
\begin{align}
	\label{Eq.FlightDyn.5}
	\bm F_j^B & = c_F \Omega_j^2 \bm R_{Rj}^B \bm e_3, ~ j = 1, \cdots, 6 \\
	\label{Eq.FlightDyn.6}
	\bm \tau_j^B & = (-1)^{j+1} c_\tau \Omega_j^2 \bm R_{Rj}^B \bm e_3, ~ j = 1, \cdots, 6
\end{align}
where $c_F$ and $c_\tau$ are constants that link the rotational speed $\Omega_j$ of the propeller to the generated thrust and torque.

Utilizing \eqref{Eq.FlightDyn.5} and \eqref{Eq.FlightDyn.6}, the control wrench $\bm W_c$ can be expressed as
\begin{equation}
	\begin{aligned}
		\label{Eq.FlightDyn.7}
		\bm W_c = \left[
		\begin{array}{c}
			\sum_{j=1}^6 \bm F_j^B \\
			\sum_{j=1}^6 \left[\bm p_{Rj}^B \right]_\times \bm F_j^B + \bm \tau_j^B
		\end{array}
		\right]
		:= \bm A \bm \Omega
	\end{aligned}
\end{equation}
where $\bm A \in \mathbb{R}^{6 \times 6}$ is an allocation matrix including the geometric and physical characteristics of the aerial vehicle;
the vector $\bm \Omega = [\Omega_1^2, \cdots, \Omega_6^2]^\top \in \mathbb{R}^6$.

\section{Composite Control}

As depicted in Fig. \ref{fig_controller}, this section presents the composite control framework of the AeRSoM robot for precise end-effector trajectory tracking, including the robust dynamic flight controller and RL-based adaptive soft robotic arm controller.
The flight controller is designed to track the desired pose trajectory under uncertainties and disturbances.
The soft robotic arm controller is employed to track the desired end position under external loads and interferences.

\subsection{Robust Dynamic Flight Control}

This subsection focuses on developing a pose trajectory tracking controller for a fully-actuated aerial vehicle, taking into consideration unmodeled dynamics and external disturbances.
To establish the dynamic coupling caused by the motion of the manipulator, the linear and angular momentum theorem  can be employed to model the term $\bm W_m = [\bm F_m^\top, \bm \tau_m^\top]^\top$ in the body frame $\mathcal{F}_B$ as \cite{2020ZhangTIE, 2024WangTRO}
\begin{equation}
	\begin{aligned}
		\label{Eq.FlightControl.0}
		\bm F_m = & m_m [\bm \omega_b]_\times ([\bm \omega_b]_\times \bm p_{cm}^b) + m_m [\dot{\bm \omega}_b]_\times \bm p_{cm}^b \\
		& + 2 m_m [\bm \omega_b]_\times \dot{\bm p}_{cm}^b + m_m \ddot{\bm p}_{cm}^b \\
		\bm \tau_m = & - \bm J_m \dot{\bm \omega}_b - [\bm \omega_b]_\times \bm J_m \bm \omega_b + m_m [\bm p_{cm}^b]_\times \bm R_I^B (g \bm e_3 - \ddot{\bm p}_b^I) \\
		& - \dot{\bm J}_m \bm \omega_b - m_m [\bm \omega_b]_\times ([\bm p_{cm}^b]_\times \dot{\bm p}_{cm}^b) - m_m [\bm p_{cm}^b]_\times \ddot{\bm p}_{cm}^b
	\end{aligned}
\end{equation}
where $m_m \in \mathbb{R}$ and $\bm J_m \in \mathbb{R}^{3 \times 3}$ are the mass and inertial matrix of the manipulator, respectively;
$\ddot{\bm p}_b^I \in \mathbb{R}^3$ is the acceleration of the flight platform expressed in the inertial frame $\mathcal{F}_I$;
$\bm p_{cm}^b$ is the position of the center of mass of the manipulator in the body frame $\mathcal{F}_B$.

Then, in order to address all uncertainties and external disturbances acting on the aerial vehicle, the generalized momentum-based external wrench estimator is introduced to enhance the robustness of the system \cite{2017TomicTRO, 2024LiangTII}, which is expressed as
\begin{equation}
	\begin{aligned}
		\label{Eq.FlightControl.1}
		\hat{\bm W}_e = \bm K_e \left[ \bm M \bm t_b - \int \left( \bm W_c - \bm C \bm t_b - \bm G + \bm W_m + \hat{\bm W}_e \right) dt \right]
	\end{aligned}
\end{equation}
where $\bm K_e \in \mathbb{R}^{6 \times 6}$ denotes the estimator positive definite gain, and the vector $\hat{\bm W}_e$ is the estimated value of the $\bm W_e$.
Differentiating \eqref{Eq.FlightControl.1}, it gives as
\begin{equation}
	\begin{aligned}
		\label{Eq.FlightControl.2}
		\dot{\hat{\bm W}}_e = \bm K_e (\bm W_e - \hat{\bm W}_e)
	\end{aligned}
\end{equation}
which implies that the estimated value $\hat{\bm W}_e$ follows the external wrench $\bm W_e$ through the first-order low-pass filter.

Based on the wrench estimator, the feedback linearization method is employed to compute the control wrench $\bm W_c$ to control the pose of the aerial vehicle as
\begin{equation}
	\begin{aligned}
		\label{Eq.FlightControl.3}
		\bm W_c = \bm M \dot{\bm t}_b^\star + \bm C \bm t_b + \bm G - \bm W_m - \hat{\bm W}_e
	\end{aligned}
\end{equation}
where the term $\dot{\bm t}_b^\star \in \mathbb{R}^6$ is a virtual control input that implements a PD control behavior to track the pose of the aerial vehicle, and it gives as
\begin{equation}
	\begin{aligned}
		\label{Eq.FlightControl.4}
		\dot{\bm t}_b^\star = \dot{\bm t}_{b,d} - \bm K_p \bm e_p - \bm K_v \bm e_v
	\end{aligned}
\end{equation}
where $\dot{\bm t}_{b,d} \in \mathbb{R}^6$ is the desired acceleration term of the aerial vehicle, and $\bm K_p$, $\bm K_v \in \mathbb{R}^{6 \times 6}$ are the PD positive definite gains.
The vectors $\bm e_p \in \mathbb{R}^6$ and $\bm e_v \in \mathbb{R}^6$ denote the pose and twist tracking errors expressed in the body frame $\mathcal{F}_B$, respectively.
The tracking errors are defined as
\begin{equation}
	\begin{aligned}
		\label{Eq.FlightControl.5}
		\bm e_p & = \left[
		\begin{array}{c}
			\bm R_I^B (\bm p_b^I - \bm p_{b,d}^I) \\
			\frac{1}{2} (\bm R_{I,d}^B \bm R_B^I - \bm R_I^B \bm R_{B,d}^I)^\vee \\
		\end{array}
		\right] \\
		\bm e_v & = \left[
		\begin{array}{c}
			\bm R_I^B (\bm v_b^I - \bm v_{b,d}^I) \\
			\bm \omega_b - \bm R_I^B \bm R_{B,d}^I \bm \omega_{b,d} \\
		\end{array}
		\right]
	\end{aligned}
\end{equation}
where $\bm p_{b,d}^I \in \mathbb{R}^3$ and $\bm R_{B,d}^I \in SO(3)$ are the desired pose of the aerial vehicle, where $\bm R_{I,d}^B$ is the transpose of $\bm R_{B,d}^I$.
The vectors $\bm v_I \in \mathbb{R}^3$ and $\bm \omega_I \in \mathbb{R}^3$ are the desired twist of the aerial vehicle in the body frame $\mathcal{F}_I$.
The symbol $(\cdot)^\vee: SO(3) \rightarrow \mathbb{R}^3$ is the inverse of the operator $[\cdot]_\times$.

\subsection{Learning-Assisted Adaptive Control of Soft Robotic Arm}


Accurate control of pneumatic soft manipulators is challenging due to hysteresis, payload variations, and interaction-dependent deformation.
To address the dominant hysteresis nonlinearity, the static pressure-length prediction model can be adopted as a feedforward compensation module.
While the model significantly improves the pressure–length mapping accuracy, residual tracking errors remain under varying operating conditions.
These residual errors are mainly caused by factors that are difficult to model explicitly, including payload changes, gravity-induced deformation, pneumatic parameter variations, and external interaction disturbances.

To improve robustness against these uncertainties, an online adaptive compensation mechanism is incorporated into the control framework.
The adaptive component continuously adjusts controller parameters according to the observed tracking behavior, thereby enhancing tracking performance under varying manipulation conditions.

\subsubsection{Hysteresis Model}

\begin{figure}
	\centering
	\includegraphics[width=3.1in]{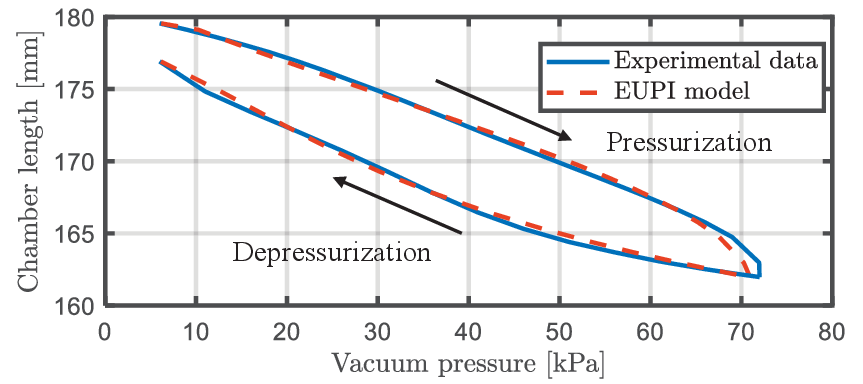}
	\caption{Pressure-length hysteresis curves using the EUPI model.}
	\label{fig_EUPImodel}
\end{figure}

To characterize the relationship between chamber length and pneumatic pressure, an isotonic test is carried out under unloaded conditions.
Due to the elasticity of the material, each actuator chamber exhibits asymmetric hysteresis characteristics.
To describe this hysteresis phenomenon, the extended unparallel Prandtl-Ishlinskii (EUPI) model is employed and expressed as \cite{2015SunICRA}
\begin{equation}
	\begin{aligned}
		\label{Eq.ArmControl.1}
		\left\{
		\begin{array}{l}
			u_p^{(k)} = \Gamma_{\mathrm{CPI}}\left(l^{(k)}\right) + \Gamma_{\mathrm{UPI}}\left(l^{(k)}\right) + W\left(l^{(k)}\right) \vspace{1ex} \\
			\Gamma_{\mathrm{CPI}}\left(l^{(k)}\right) = a_0 l^{(k)} + \sum_{j=1}^{N_c} b_j G_{\gamma_j,c_j,1}\left(l^{(k)}\right) \vspace{1ex} \\
			\Gamma_{\mathrm{UPI}}\left(l^{(k)}\right) = \sum_{j=1}^{N_u} \delta_j G_{\gamma_j,c_j,d_j}\left(l^{(k)}\right) \vspace{1ex} \\
			W\left(l^{(k)}\right) = \sum_{j=2}^{N_w} w_j (l^{(k)})^j +w_0 \vspace{1ex} \\
			G_{\gamma_j,c_j,d_j}\left(l^{(k)}\right) = \max\{c_j(l^{(k)}-\gamma_j), \vspace{1ex} \\
			~~~~~~~~~~~~~~ \min\{d_j(l^{(k)}+\gamma_j), G_{\gamma_j,c_j,d_j}\left(l^{(k-1)}\right)\} \}
		\end{array}
		\right.
	\end{aligned}
\end{equation}
where $l^{(k)}$ is the chamber length and $u_p^{(k)}$ represents the actuated chamber pressure predicted by the EUPI model, which consists of the symmetric portion $\Gamma_{\mathrm{CPI}}\left(l^{(k)}\right)$, the asymmetric portion $\Gamma_{\mathrm{UPI}}\left(l^{(k)}\right)$, and the polynomial portion $W\left(l^{(k)}\right)$;
$G_{\gamma_j,c_j,d_j}\left(l^{(k)}\right)$ is the operator output of the unparallel Prandtl-Ishlinskii model;
$a_0$ is the linear weight gain for amplifying $l^{(k)}$, and $b_i$, $\delta_{ij}$, $w_i$ are the weight gains;
$\gamma_j$ is the $j$-th dead zone, and $c_j$, $d_j$ are the $j$-th tilted angles of the pressurization and depressurization edges, respectively;
$w_0$ is an offset associated with hysteresis loops working angle;
$N_c$ and $N_w$ are the numbers in the symmetric and polynomial portions, respectively;
$N_u$ is the total number of the dead zone in the asymmetric portion.
The EUPI fitting curve is plotted in Fig. \ref{fig_EUPImodel}.

\subsubsection{Learning-Assisted Adaptive Control Design}

The pressure command $u_a^{(k)}$ applied to the soft actuator is composed of a feedforward compensation term $u_p^{(k)}$ generated by the EUPI model and a feedback correction term $u_c^{(k)}$:
	\begin{align}
		\label{Eq.ArmControl.2}
		u_a^{(k)} = & u_p^{(k)} + u_c^{(k)} \\
		\label{Eq.ArmControl.3}
		u_c^{(k)} = & k_p \tilde{l}^{(k)} + k_i \int \tilde{l}^{(k)} dt + k_d \dot{\tilde{l}}^{(k)}
	\end{align}
where $\tilde{l}^{(k)} = l_d^{(k)} - l^{(k)}$ with the desired chamber length $l_d^{(k)}$ and the actual chamber length $l^{(k)}$, which can be calculated by the desired position $\bm p_{S0,d}^{S1}$ and actual position $\bm p_{S0}^{S1}$ according to the inverse kinematic model \eqref{Eq.IKArm.1}-\eqref{Eq.IKArm.3}.
The parameters $k_p$, $k_i$, and $k_d$ are the positive proportional, integral, and derivative coefficients, respectively.
To improve the feedback control performance, the proportional coefficient $k_p$ is adjusted and set as an exponential function, which is expressed as
\begin{equation}
	\begin{aligned}
		\label{Eq.ArmControl.4}
		k_p = k_{p0} + \lambda_1 e^{( \lambda_2 - \frac{\lambda_3}{|\tilde{l}^{(k)}|})}
	\end{aligned}
\end{equation}
where $k_{p0}$, $\lambda_1$, $\lambda_2$, and $\lambda_3$ are positive constants.
To facilitate the smooth adjustment of the coefficient $k_p$, the Sarsa learning algorithm is employed to determine the parameters online.
To assess the end tracking performance of the soft robotic arm, the state space $\mathbb{S}$ is defined and divided into several continuous and symmetric intervals as
\begin{equation}
	\begin{aligned}
		\label{Eq.ArmControl.5}
		& \mathbb{S} = \left\{ \bm S_1, ~\bm S_2, ~\bm S_3, ~\bm S_4, ~\bm S_5, ~\bm S_6, ~\bm S_7 \right\} \\
		& \left\{
		\begin{array}{ll}
			\bm S_1: \tilde{l} \in (-\infty, -8); & \bm S_2: \tilde{l} \in [-8, -2); \vspace{0.5ex} \\
			\bm S_3: \tilde{l} \in [-2, -0.5); & \bm S_4: \tilde{l} \in [-0.5, 0.5]; \vspace{0.5ex} \\
			\bm S_5: \tilde{l} \in (0.5, 2]; & \bm S_6: \tilde{l} \in (2, 8]; \vspace{0.5ex} \\
			\bm S_7: \tilde{l} \in (8, +\infty).
		\end{array}
		\right.
	\end{aligned}
\end{equation}

Then, an action space $\mathbb{A}$ containing four actions is set to adjust the parameters $k_{p0}, \lambda_1, \lambda_2, \lambda_3$, and it is expressed as
\begin{equation}
	\begin{aligned}
		\label{Eq.ArmControl.6}
		& \mathbb{A} = \left\{ A_1, ~A_2, ~A_3, ~A_4 \right\} \\
		& A_i: [k_{p0} ~\lambda_1 ~\lambda_2 ~\lambda_3], ~ i = 1, \cdots, 4
	\end{aligned}
\end{equation}
It indicates that at the current state $\tilde{l}^{(k)}$, an action $A^{(k)}$ can be chosen from the action space $\mathbb{A}$ to determine the proportional coefficient $k_p$ by \eqref{Eq.ArmControl.6}.
Further, to evaluate the chosen action, a reward table $\bm R(\mathbb{S}, \mathbb{A}) \in \mathbb{R}^{7 \times 4}$ is designed based on the state space $\mathbb{S}$ and the action space $\mathbb{A}$.

Next, to enable the soft robotic arm to learn to choose the best action, the improved $\epsilon$-greedy policy is used to reduce the variety of action choices and improve the convergence rate.
The modified $\epsilon$-greedy policy is given as
\begin{equation}
	\begin{aligned}
		\label{Eq.ArmControl.7}
		\left\{
		\begin{array}{l}
			\hbox{if} ~~\mathrm{rand}() < \epsilon, ~~A^{(k)} \leftarrow \mathrm{rand}_A (\mathbb{A}\left\{ A_1, ~A_2, ~A_3, ~A_4 \right\}) \vspace{0.5ex} \\
			\hbox{else if} ~~\tilde{l}^{(k)} \in \bm S_1, ~~~A^{(k)} \leftarrow \max_A \bm V(\bm S_1, \{A_1\}) \vspace{0.5ex} \\
			\hbox{else if} ~~\tilde{l}^{(k)} \in \bm S_2, ~~~A^{(k)} \leftarrow \max_A \bm V(\bm S_2, \{A_1, ~A_2\}) \vspace{0.5ex} \\
			\hbox{else if} ~~\tilde{l}^{(k)} \in \bm S_3, ~~~A^{(k)} \leftarrow \max_A \bm V(\bm S_3, \{A_2, ~A_3\}) \vspace{0.5ex} \\
			\hbox{else if} ~~\tilde{l}^{(k)} \in \bm S_4, ~~~A^{(k)} \leftarrow \max_A \bm V(\bm S_4, \{A_3, ~A_4\}) \vspace{0.5ex} \\
			\hbox{else if} ~~\tilde{l}^{(k)} \in \bm S_5, ~~~A^{(k)} \leftarrow \max_A \bm V(\bm S_5, \{A_2, ~A_3\}) \vspace{0.5ex} \\
			\hbox{else if} ~~\tilde{l}^{(k)} \in \bm S_6, ~~~A^{(k)} \leftarrow \max_A \bm V(\bm S_6, \{A_1, ~A_2\}) \vspace{0.5ex} \\
			\hbox{else if} ~~\tilde{l}^{(k)} \in \bm S_7, ~~~A^{(k)} \leftarrow \max_A \bm V(\bm S_7, \{A_1\})
		\end{array}
		\right.
	\end{aligned}
\end{equation}
where $\epsilon \in (0, 1)$ and $\bm V(\mathbb{S}, \mathbb{A}) \in \mathbb{R}^{7 \times 4}$ is the state-action value, which is formulated by
\begin{equation}
	\begin{aligned}
		\label{Eq.ArmControl.8}
		\bm V(\bm S_i^{(k)}, A_i^{(k)}) \leftarrow \bm V(\bm S_i^{(k)}, A_i^{(k)}) + \beta [ \bm R(\bm S_i^{(k+1)}, A_i^{(k+1)}) \\
		+ \sigma \bm V(\bm S_i^{(k+1)}, A_i^{(k+1)}) - \bm V(\bm S_i^{(k)}, A_i^{(k)})]
	\end{aligned}
\end{equation}
where $\beta$ is the learning rate, and $\sigma$ denotes the discount factor.
Following \eqref{Eq.ArmControl.2}-\eqref{Eq.ArmControl.8}, the learning-assisted adaptive controller of the soft robotic arm can be realized.
The corresponding algorithm is illustrated in Algorithm \ref{alg_1}.

\subsubsection{Stiffening Layer Response}

During free flight, the soft robotic arm is expected to achieve higher stiffness for precise end trajectory tracking.
In contrast, lower stiffness allows the soft arm to perform various manipulation tasks with greater compliance.
To minimize computational cost and complexity, the stiffening layer is activated using an ON-OFF response, which is expressed as
\begin{equation}
	\begin{aligned}
		\label{Eq.StiffeningLayer.1}
		u_s^{(k)} = \left\{
		\begin{array}{ll}
			\hbox{ON}, & \hbox{Jamming;} \\
			\hbox{OFF}, & \hbox{Unjamming.}
		\end{array}
		\right.
	\end{aligned}
\end{equation}

\begin{remark}
	\label{remark1}
	Regarding the rotary mechanism, the servo angle is commanded by the inner PID position controller.
	In particular, the nested velocity and acceleration profiles are regulated to ensure smooth and steady motion of the rotary mechanism.
\end{remark}

\subsection{End-Effector Trajectory Generation}

\begin{figure}
	\centering
	\includegraphics[width=3.3in]{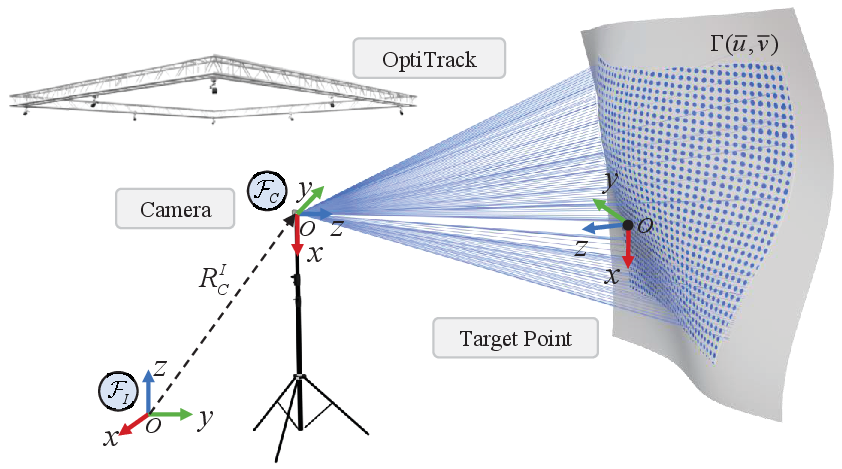}
	\caption{Offline measurement of task target point, which is constructed the desired end-effector trajectory.}
	\label{fig_pointCloud}
\end{figure}

For the task target point measurement, the target point is obtained offline using the Intel RealSense D435i depth camera, as shown in Fig. \ref{fig_pointCloud}.
The camera generates a depth image, which is then converted into point cloud information.
The point cloud information is filtered and sparsely processed to filter out noise and prevent overfitting.
Then, the least squares method is used to fit the processed point cloud information to obtain a nonuniform rational B-splines (NURBS) surface, which can be represented as \cite{2017Abbena}
\begin{equation}
	\begin{aligned}
		\label{Eq.PointCloud.1}
		\bm p_{e,d}^c = \bm \Gamma(\bar{u}, \bar{v}) = \frac{\sum_{i=0}^{m} \sum_{j=0}^{n} N_i^{\bar{p}} (\bar{u}) M_j^{\bar{q}} (\bar{v}) \bar{w}_{ij} \bm P_{ij}}{\sum_{i=0}^{m} \sum_{j=0}^{n} N_i^{\bar{p}}(\bar{u}) M_j^{\bar{q}} (\bar{v}) \bar{w}_{ij}}
	\end{aligned}
\end{equation}
where $\bm p_{e,d}^c \in \mathbb{R}^3$ is the position of the target point expressed in the camera frame $\mathcal{F}_C$;
$\bm \Gamma(\bar{u}, \bar{v})$ is the mapping from the parametric space $\mathcal{D} = \{ (\bar{u}, \bar{v}) | \bar{u}, \bar{v} \in [0, 1] \} \in \mathbb{R}^2$ to the configuration space $\mathcal{C}$, where each $(\bar{u}, \bar{v})$ corresponds to the 3D spatial coordinate point $\bm p_e^c$ on the surface;
$\bm P_{ij}$ is the control point of the surface, which is used to define the surface shape;
$N_i^{\bar{p}} (\bar{u})$ and $M_j^{\bar{q}} (\bar{v})$ are B-spline basis functions in the directions of parameters $\bar{u}$ and $\bar{v}$, respectively, where $\bar{p}$ and $\bar{q}$ are the orders;
the parameter $\bar{w}_{ij}$ represents the weight, which affects the effort of each control point to the surface shape.

Further, based on \eqref{Eq.PointCloud.1}, the normal vector $\vec{\bm n}^c \in \mathbb{R}^3$ of the NURBS surface expressed in the camera frame $\mathcal{F}_C$ can be given as
\begin{equation}
	\begin{aligned}
		\label{Eq.PointCloud.2}
		\vec{\bm n}^c = \left. \frac{\partial \bm \Gamma(\bar{u}, \bar{v})}{\partial \bar{u}} \right|_{(\bar{u}_0, \bar{v}_0)} \times \left. \frac{\partial \bm \Gamma(\bar{u}, \bar{v})}{\partial \bar{v}} \right|_{(\bar{u}_0, \bar{v}_0)}
	\end{aligned}
\end{equation}

Through homogeneous transformation, the orientation of the target point is transformed from the camera frame $\mathcal{F}_C$ to the inertial frame $\mathcal{F}_I$, and it yields
\begin{align}
	\label{Eq.PointCloud.3}
	\bm p_{e,d}^I & = \bm p_{c,d}^I + \bm R_C^I \bm \Gamma(\bar{u}, \bar{v}) \\
	\label{Eq.PointCloud.4}
	\vec{\bm n}_d^I & = \bm R_C^I \vec{\bm n}^c
\end{align}
where $\bm p_{e,d}^I \in \mathbb{R}^3$ and $\bm p_{c,d}^I \in \mathbb{R}^3$ are the position of the target point and the camera expressed in the inertial frame $\mathcal{F}_I$, respectively;
$\bm R_C^I \in \mathbb{R}^{3 \times 3}$ is the rotation matrix from the camera frame $\mathcal{F}_C$ to the inertial frame $\mathcal{F}_I$;
$\vec{\bm n}_d^I \in \mathbb{R}^3$ is the normal vector of the target point expressed in the frame $\mathcal{F}_I$.
The resulting vectors $\bm p_{e,d}^I$ and $\vec{\bm n}_d^I$ obtained offline are used to construct the desired end-effector trajectory of the AeRSoM robot, particularly for tasks involving physical interaction with unstructured surfaces.

It is worth noting that, given the desired end-effector pose, there exist multiple ways to realize the combined motion of the aerial vehicle, rotary mechanism, and soft robotic arm.
To solve the kinematic redundancy, the successive null-space projection method is employed to obtain the unique inverse kinematics solution \cite{2008Siciliano, 2015DietrichIJRR}.
The successive equations are given as
\begin{align}
	\label{Eq.SNP.1}
	\mathbf{\dot{q}}_i & = \mathbf{\dot{q}}_{i-1} + (\mathbf{J}_i \mathbf{N}_{i-1})^\dagger (\mathbf{\dot{x}}_{i,d} - \mathbf{J}_i \mathbf{\dot{q}}_{i-1}) \\
	\label{Eq.SNP.2}
	\mathbf{N}_i & = \mathbf{N}_{i-1} - (\mathbf{J}_i \mathbf{N}_{i-1})^\dagger (\mathbf{J}_i \mathbf{N}_{i-1})
\end{align}
where $\mathbf{\dot{q}}_0 = \mathbf{0}$ and $\mathbf{N}_0 = \mathbf{I}$;
the final instruction $\mathbf{\dot{q}}_i$ ensures that high-priority tasks are completed, while utilizing system redundancy to implement low-priority tasks in the null space.

Therefore, in this work, we define the end-effector pose as the highest priority, followed by the attitude stabilization of the aerial base, then the joint of the one-DOF rotary mechanism, and finally the position constraints of the soft robotic arm within its reachable workspace.
Define the desired configuration variables of the AeRSoM robot $\mathbf{q}_d = [(\bm p_{b,d}^I)^\top, ((\bm R_{B,d}^I)^\vee)^\top, \eta_{m,d}, (\bm p_{S0,d}^{S1})^\top]^\top$, where $\eta_{m,d}$ is the desired joint angle of the rotary mechanism.
Based on \eqref{Eq.SNP.1} and \eqref{Eq.SNP.2}, we can compute $\mathbf{\dot{q}}_d$ as
\begin{equation}
	\begin{aligned}
		\label{Eq.SNP.3}
		\mathbf{\dot{q}}_d = & \mathbf{J}_1^\dagger \mathbf{\dot{x}}_{1,d} + (\mathbf{I} - \mathbf{J}_1^\dagger \mathbf{J}_1) [\mathbf{J}_2^\dagger \mathbf{\dot{x}}_{2,d} \\
		& + \mathbf{N}_2 \mathbf{J}_3^\dagger \mathbf{\dot{x}}_{3,d} + \mathbf{N}_3 \mathbf{J}_4^\dagger \mathbf{\dot{x}}_{4,d}]
	\end{aligned}
\end{equation}
where $\mathbf{J}_i$, $i = 1, \cdots, 4$ represents the Jacobian matrix of the $i$th task;
$\mathbf{J}_i^\dagger = \mathbf{B}_i^{-1} \mathbf{J}_i^\top (\mathbf{J}_i \mathbf{B}_i^{-1} \mathbf{J}_i^\top)^{-1}$ with the invertible weight matrix $\mathbf{B}_i$.
$\mathbf{x}_{1,d}$ is given by $\bm p_{e,d}^I$ and $\vec{\bm n}_d^I$.
For $\mathbf{x}_{2,d}$, we regulate the desired roll and pitch angles as zero to ensure stable attitude of the aerial vehicle, and the desired yaw angle is also set as zero in this work.
The specified angle $\eta_{m,d}$ is employed for $\mathbf{x}_{3,d}$.
Finally, $\mathbf{x}_{4,d}$ specifies that the end position of the soft robotic arm is within its reachable workspace.

\section{Real-World Experiments}

\begin{figure}
	\centering
	\includegraphics[width=3.3in]{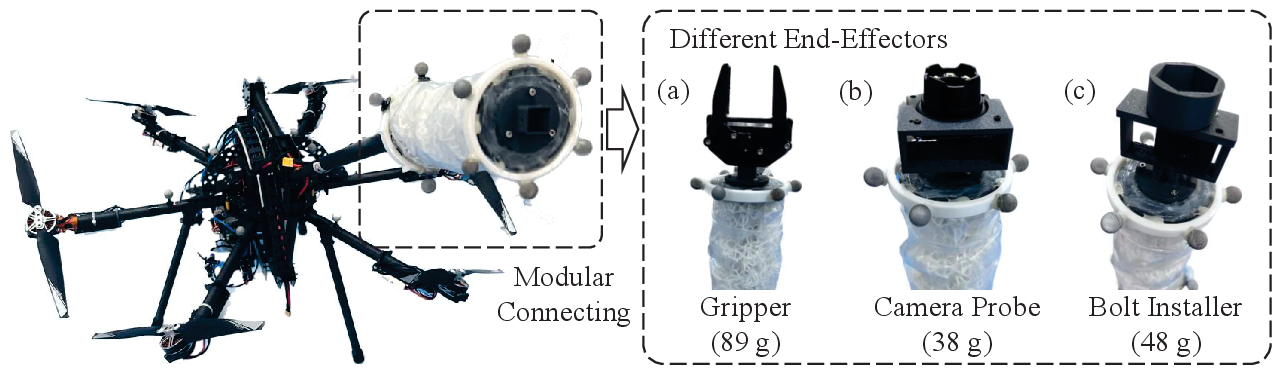}
	\caption{AeRSoM robot equipped with different end-effectors ((a) gripper, (b) camera probe, and (c) bolt installer) for various aerial manipulation tasks.
		The end-effectors can be connected modularly and altered rapidly.}
	\label{fig_endEffector}
\end{figure}

This section highlights the capabilities and applications of the AeRSoM robot through a series of real-world experiments.
The tracking control performance of both the aerial platform and the soft robotic arm is verified to provide a basis for aerial manipulation.
As depicted in Fig. \ref{fig_endEffector}, the AeRSoM robot is equipped with various lightweight end-effectors to carry out complex aerial manipulation tasks, including dynamic transmission-line grasping, physical interaction with a wind turbine blade, peg-in-hole, and screwing.
In particular, the end-effectors can be connected modularly and altered rapidly.

\begin{table}[h]
	\centering
	\renewcommand{\arraystretch}{1.2}
	\caption{Selection of the controller gains and parameters
		\label{Table_Parameter}}
	\begin{tabular}{ll}
		\toprule
		Parameter & Value \\
		\midrule
		$\bm M^\ast$ & $\mathrm{diag} (8.875 \bm{I}_{3}, 0.1560, 0.1568, 0.2908)$ \\
		$\bm K_p$ & $\bm M^{-1} \mathrm{diag}(20 \bm{I}_{3}, 0.3 \bm{I}_{3})$ \\
		$\bm K_v$ & $\bm M^{-1} \mathrm{diag}(12 \bm{I}_{3}, 0.2 \bm{I}_{3})$ \\
		$\bm K_e$ & $\mathrm{diag}(1.2 \bm{I}_{3}, 0.25 \bm{I}_{3})$ \\
		$k_i$ & $0.15$ \\
		$k_d$ & $0.5$ \\
		$m_m$ & $2.497$ \\
		$\bm J_m$ & $\mathrm{diag} (0.0276, 0.1150, 0.1418)$ \\
		\bottomrule
	\end{tabular}\\[10pt]
	\begin{tablenotes}
		\footnotesize
		\item[] $^\ast$The mass is measured by an electronic scale, and the primary components of inertia are obtained from the CAD model.
	\end{tablenotes}
\end{table}

\subsection{Experimental Settings}

The gains and parameters for the robust flight controller and the adaptive soft robotic arm controller are chosen and listed in Table \ref{Table_Parameter}, where the parameter $k_p$ is obtained based on the Sarsa learning algorithm.
The overall system architecture is depicted in Fig. \ref{fig_system}(c).
The pose signal is obtained through the OptiTrack motion tracking system, which transmits data to the onboard computer via a wireless router.
The end-effectors, including the gripper and bolt installer, are driven by the Dynamixel XC330-T288-T servo.
Additionally, the camera probe connects with the onboard computer through a USB interface.
All the software for the proposed AeRSoM robot is developed using ROS Noetic on Ubuntu 20.04.
The robust flight controller runs at 200 Hz, while the adaptive soft robotic arm controller runs at 20 Hz.

\subsection{Robust Flight Test}

\begin{figure}
	\centering
	\includegraphics[width=3.3in]{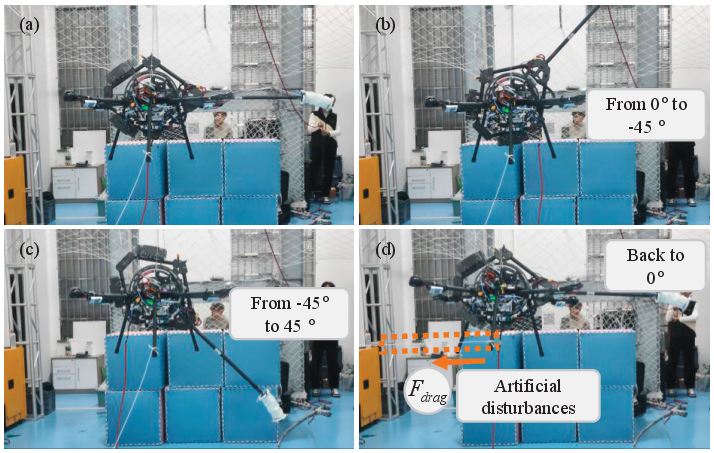}
	\caption{Snapshots of robust flight tests, where the rotary mechanism rotates to different angles with 0.48 rad/s (from a) 0$^\circ$ to b) -45$^\circ$ to c) 45$^\circ$ back to d) 0$^\circ$) and then the aerial platform is imposed on random artificial disturbances.}
	\label{fig_Exp1_Snapshot}
\end{figure}

\begin{figure}
	\centering
	\includegraphics[width=3.3in]{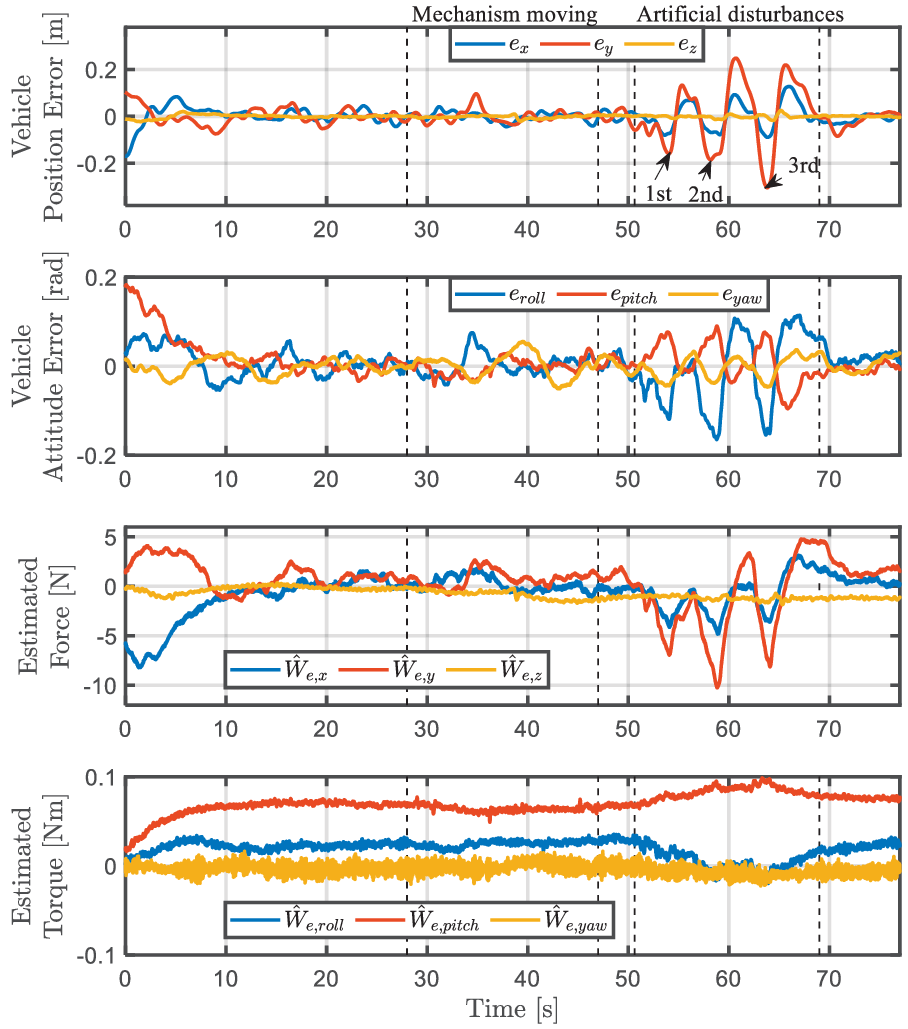}
	\caption{Experimental results of pose errors and estimated wrenches of the aerial vehicle under the coupling disturbances from the rotary mechanism and artificial disturbances.}
	\label{fig_Exp1}
\end{figure}

\begin{table}[h]
	\centering
	\renewcommand{\arraystretch}{1.2}
	\caption{Maximum $\kappa$, mean $\mu$, and standard deviation $\sigma$ of the absolute errors (Units: cm, deg)
		\label{Table_Exp1}}
	\begin{tabular}{cc|c|c|c|c|c|c}
		\toprule
		&     & $e_x$ & $e_y$ & $e_z$ & $e_{roll}$ & $e_{pitch}$ & $e_{yaw}$ \\ \hline
		\multicolumn{1}{c|}{\multirow{5}{*}{$\kappa$}} & I   & 16.98 & 30.43 & 2.52  & 9.44       & 10.46       & 3.17      \\ \cline{2-8}
		\multicolumn{1}{c|}{}                          & II  & 16.98 & 10.12 & 2.27  & 4.16       & 10.46       & 2.24      \\ \cline{2-8}
		\multicolumn{1}{c|}{}                          & III & 4.38  & 9.65  & 2.04  & 4.30       & 2.57        & 3.17      \\ \cline{2-8}
		\multicolumn{1}{c|}{}                          & IV  & 12.75 & 30.43 & 2.52  & 9.44       & 5.47        & 2.69      \\ \cline{2-8}
		\multicolumn{1}{c|}{}                          & V   & 1.84  & 2.20  & 0.49  & 1.97       & 1.04        & 1.69      \\ \hline
		\multicolumn{1}{c|}{\multirow{5}{*}{$\mu$}}    & I   & 2.77  & 4.74  & 0.48  & 1.96       & 1.51        & 1.00      \\ \cline{2-8}
		\multicolumn{1}{c|}{}                          & II  & 2.67  & 2.77  & 0.60  & 1.53       & 1.99        & 0.81      \\ \cline{2-8}
		\multicolumn{1}{c|}{}                          & III & 1.51  & 1.90  & 0.45  & 1.02       & 0.80        & 1.22      \\ \cline{2-8}
		\multicolumn{1}{c|}{}                          & IV  & 4.91  & 12.12 & 0.47  & 4.15       & 2.12        & 1.18      \\ \cline{2-8}
		\multicolumn{1}{c|}{}                          & V   & 0.81  & 1.08  & 0.14  & 1.25       & 0.51        & 0.90      \\ \hline
		\multicolumn{1}{c|}{\multirow{5}{*}{$\sigma$}} & I   & 2.75  & 5.82  & 0.48  & 1.94       & 1.92        & 0.73      \\ \cline{2-8}
		\multicolumn{1}{c|}{}                          & II  & 2.81  & 2.27  & 0.57  & 1.16       & 2.64        & 0.61      \\ \cline{2-8}
		\multicolumn{1}{c|}{}                          & III & 1.10  & 1.83  & 0.40  & 0.87       & 0.61        & 0.91      \\ \cline{2-8}
		\multicolumn{1}{c|}{}                          & IV  & 3.28  & 7.45  & 0.47  & 2.45       & 1.57        & 0.72      \\ \cline{2-8}
		\multicolumn{1}{c|}{}                          & V   & 0.47  & 0.57  & 0.12  & 0.35       & 0.30        & 0.43      \\ \bottomrule
	\end{tabular}\\[10pt]
	\begin{tablenotes}
		\footnotesize
		\item[] I, II, III, IV, and V represent time intervals [0, 77]s, [0, 28]s, [28, 47]s, [50.6, 69]s, and [73, 77]s, respectively.
	\end{tablenotes}
\end{table}


This test demonstrates the performance of the AeRSoM during free flight, specifically in response to coupling disturbances from the rotary mechanism as well as artificial disturbances.
The key snapshots of the robust flight test are shown in Fig. \ref{fig_Exp1_Snapshot}.
Initially, the aerial vehicle is instructed to track a desired spatial point, with the rotary mechanism fixed in the horizontal state (Fig. \ref{fig_Exp1_Snapshot}(a)).
The Dynamixel servo is continuously operated at an angular velocity of 0.48 rad/s.
As a result of the transmission conversion, the rotary mechanism rotates clockwise or counterclockwise at different angles before ultimately returning to the horizontal position (Fig. \ref{fig_Exp1_Snapshot}(b)-(d)).
As shown in Fig. \ref{fig_Exp1_Snapshot}(d), the landing gear of the aerial vehicle is secured with a thin wire, which is pulled randomly three times to introduce artificial disturbances to the vehicle.
The test results are plotted in Fig. \ref{fig_Exp1}.
Specifically, the first and second graphs of Fig. \ref{fig_Exp1} display the position and attitude errors of the aerial vehicle, respectively.
The robust flight controller responds effectively, allowing the AeRSoM to converge near the target spatial point after approximately 5.5s.
From 28s to 47s, the rotary mechanism is commanded to move.
During this time, there is only a spike in the error curve at about 35s, which may result from the large amplitude movement of the rotary mechanism (from -45$^\circ$ to 45$^\circ$).
For the remainder of this period, the error remains within 4 cm under coupling disturbances.
After 50.6s, artificial disturbances are applied to the robot body by pulling the wire, causing the maximum roll inclination of 10 degrees.
After that, the position error of the vehicle converges to within 2 cm.
To further quantitatively evaluate the control performance of the aerial vehicle, three indices (maximum $\kappa$, mean $\mu$, and standard deviation $\sigma$ of the absolute errors) are listed in Table \ref{Table_Exp1}.
The outputs of the corresponding force and torque estimators are plotted in the third and fourth graphs of Fig. \ref{fig_Exp1}, and the bias in the estimated torque values is caused by model errors.
This test demonstrates that the AeRSoM robot can tolerate strong coupling of the manipulator and external disturbances while maintaining robust and accurate pose tracking.
This capacity provides a solid foundation for subsequent end-effector tracking tasks.


\subsection{Constant Curvature Model Validation}

\begin{figure}
	\centering
	\includegraphics[width=3.2in]{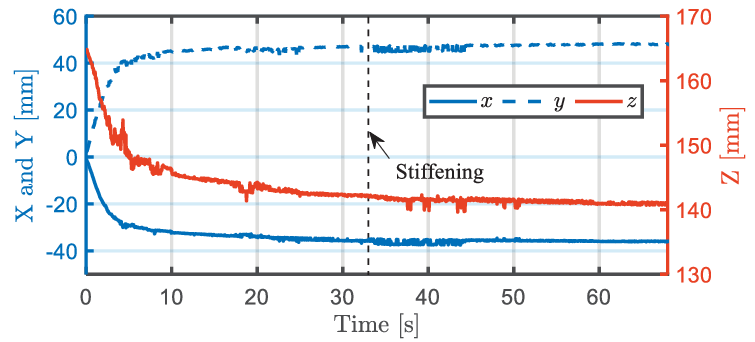}
	\caption{Motion test of the soft robotic arm from unjamming to jamming.}
	\label{fig_Exp_motionTest}
\end{figure}

\begin{figure}
	\centering
	\includegraphics[width=3.3in]{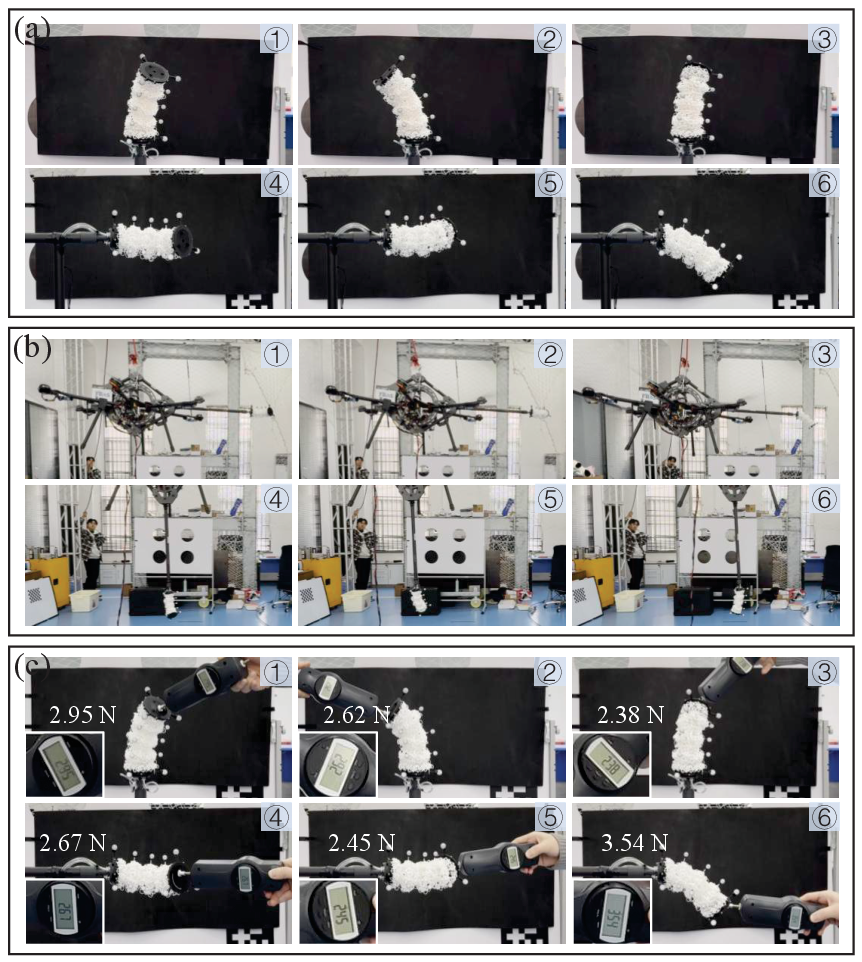}
	\caption{Snapshots of the constant curvature model validation process.
		(a) Free bending.
		(b) Consider the influence by the motion of the aerial base.
		(c) Consider the effect of the interaction with the environment, where a thrust gauge is used to apply the interaction for the soft robotic arm, and the maximum interaction force is marked in the snapshots.
		(a1)-(a3), (b4)-(b6), and (c1)-(c3) The soft robotic arm is placed vertically.
		(a4)-(a6), (b1)-(b3), and (c4)-(c6) The soft robotic arm is placed horizontally.}
	\label{fig_Exp_snapshot}
\end{figure}

We have added structural holders every 15 mm on the soft robotic arm (as shown in Fig. \ref{fig_softArm}(iii)) to maintain the characteristics of the constant curvature model as much as possible
To further test the accuracy of the constant curvature model of the soft robotic arm, the plan of the test is to mark the base plane, the end plane, and three intermediate cross-sections of the soft robotic arm, and use the OptiTrack motion capture system to measure its pose.
Then, the bending angles of the three cross-sections and the end plane relative to the base plane are compared through constant curvature model calculations.

\begin{figure}
	\centering
	\includegraphics[width=3.3in]{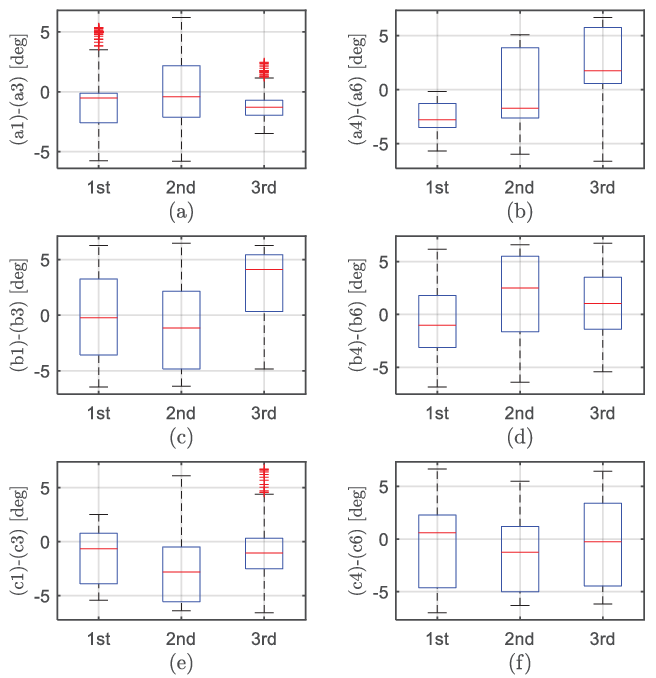}
	\caption{Bending angle errors of three cross-sections and the end face relative to the base plane based on calculations using the constant curvature model.}
	\label{fig_Exp_CCModel}
\end{figure}

To add markers to the intermediate cross-sections, the motion tests on the soft robotic arm are first conducted from unjamming to jamming, as shown in Fig. \ref{fig_Exp_motionTest}, and the results show that the motion of the soft robotic arm is almost unaffected by the unjammed stiffening layer.
Therefore, we remove the membranes from the original soft robotic arm but retain the stiffening layer structure for constant curvature model verification.
For comprehensive testing and validation, the conditions of the soft robotic arm are considered under its free bending (Fig. \ref{fig_Exp_snapshot}(a)), the influence by the motion of the aerial base (Fig. \ref{fig_Exp_snapshot}(b)), and the effect of the interaction with the environment (Fig. \ref{fig_Exp_snapshot}(c)).
In the test influenced by the motion of the aerial base, the flight trajectory is commanded as $\bm p_d = [-0.03 + 0.8 \sin(\frac{1}{8}t), 0.86 + 0.8 \cos((\frac{1}{8}t), 2.0)]^\top$.
Moreover, the testing of the soft robotic arm in both vertical and horizontal orientations is also fully taken into account.
Fig. \ref{fig_Exp_CCModel} shows the bending angle errors of three cross-sections and the end face relative to the base plane based on calculations using the constant curvature model.
To further quantify the error results, the mean absolute error is given in Table \ref{Table_BendingError}.
Small deviations in the bending angle are acceptable for the overall movement of the soft robotic arm.

\begin{table}
	\centering
	\renewcommand{\arraystretch}{1.2}
	\caption{Mean absolute error of the bending angle (Unit: deg)
		\label{Table_BendingError}}
	\begin{tabular}{c|c|c|c|c|c|c}
		\toprule
		\textbf{\begin{tabular}[c]{@{}c@{}}Cross-\\ section\end{tabular}} & \textbf{\begin{tabular}[c]{@{}c@{}}(a1)-\\ (a3)\end{tabular}} & \textbf{\begin{tabular}[c]{@{}c@{}}(a4)-\\ (a6)\end{tabular}} & \textbf{\begin{tabular}[c]{@{}c@{}}(b1)-\\ (b3)\end{tabular}} & \textbf{\begin{tabular}[c]{@{}c@{}}(b4)-\\ (b6)\end{tabular}} & \textbf{\begin{tabular}[c]{@{}c@{}}(c1)-\\ (c3)\end{tabular}} & \textbf{\begin{tabular}[c]{@{}c@{}}(c4)-\\ (c6)\end{tabular}} \\ \hline
		\textbf{1st}                                                      & 1.72                                                          & 2.55                                                          & 3.25                                                          & 2.65                                                          & 2.06                                                          & 3.88                                                          \\ \hline
		\textbf{2nd}                                                      & 2.39                                                          & 3.10                                                          & 3.37                                                          & 3.83                                                          & 3.70                                                          & 2.98                                                          \\ \hline
		\textbf{3rd}                                                      & 1.56                                                          & 3.45                                                          & 3.22                                                          & 2.63                                                          & 2.22                                                          & 3.16                                                          \\ \bottomrule
	\end{tabular}
\end{table}

\subsection{End Tracking Test of Soft Robotic Arm}

\begin{figure}
	\centering
	\includegraphics[width=3.3in]{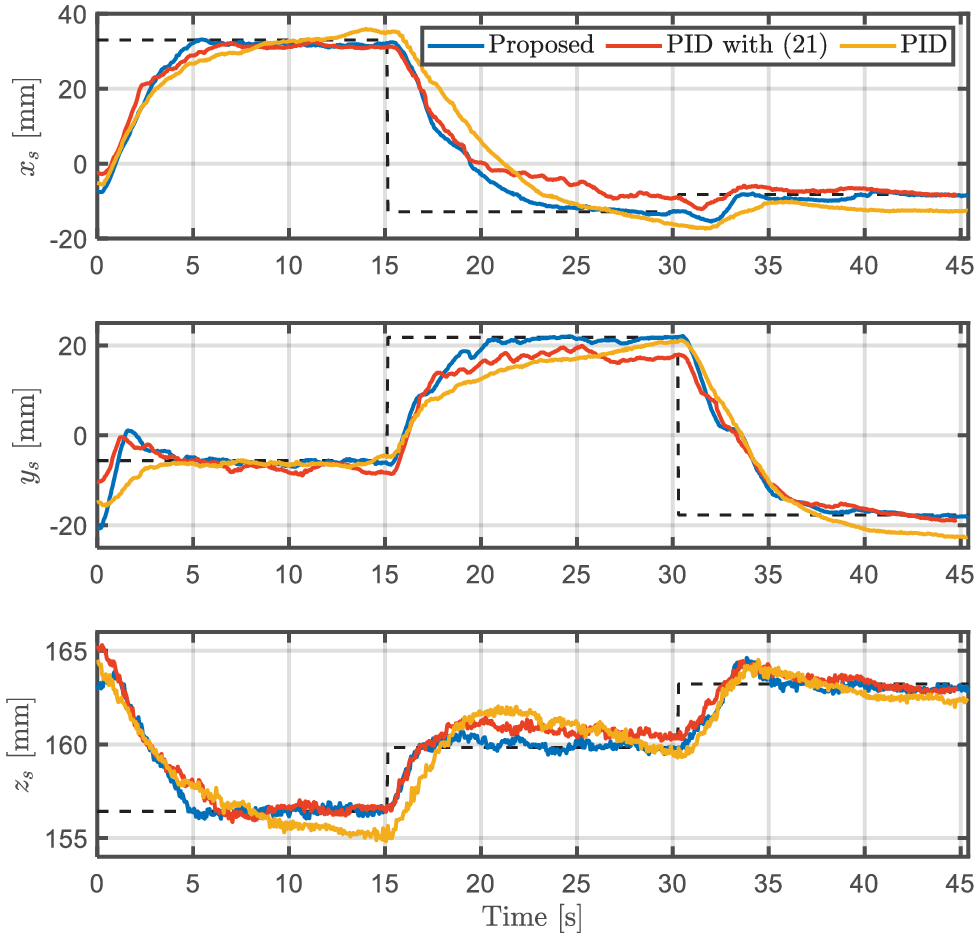}
	\caption{Experimental results of the static end tracking of the soft robotic arm via the ablation study.
		The comparisons among the proposed RL-based adaptive controller, PID with hysteresis model \eqref{Eq.ArmControl.1}, and only PID.}
	\label{fig_Exp2_1}
\end{figure}

\begin{figure}
	\centering
	\includegraphics[width=3.3in]{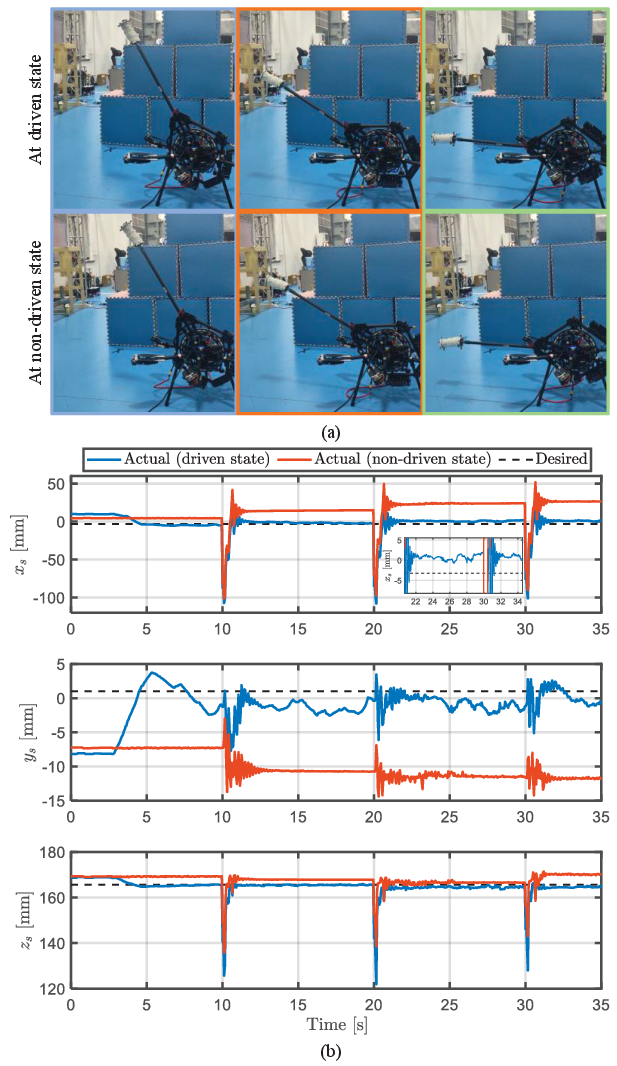}
	\caption{Experimental results of the dynamic end tracking of the soft robotic arm.
		(a) Snapshots of the soft robotic arm at the driven state or non-driven state when the rotary mechanism rotates to different angles (-60$^\circ$, -30$^\circ$, and 0$^\circ$).
		In the driven state, the vacuum pumps are active and drive the soft robotic arm using the proposed control strategy.
		Conversely, in the non-driven state, the vacuum pumps are inactive, and the soft robotic arm remains passive.
		(b) Trajectory of the end point tracking in the base frame of the soft robotic arm $\mathcal{F}_{S0}$ when the rotary mechanism is at the -90$^\circ$, -60$^\circ$, -30$^\circ$, or 0$^\circ$ state.}
	\label{fig_Exp2_2}
\end{figure}

This subsection tests the end tracking response of the soft robotic arm using the proposed learning-assisted adaptive controller.
To assess the performance of the proposed adaptive controller, a static end tracking experiment is conducted through an ablation study to track different end points within the workspace of the soft robotic arm.
In this experiment, the rotary mechanism is set and fixed at the -90$^\circ$ state, while the soft robotic arm is oriented vertically upwards.
Then, the comparative experiments are conducted to track three different points expressed in the base frame of the soft robotic arm $\mathcal{F}_{S0}$ using the proposed learning-assisted adaptive controller, a PID controller with the feedforward hysteresis model \eqref{Eq.ArmControl.1}, and a standard PID controller.
The experimental results of the static end tracking of the soft robotic arm are plotted in Fig. \ref{fig_Exp2_1}.
The tracking curves indicate that the proposed RL-based adaptive controller achieves the highest tracking accuracy and fastest convergence speed.

Further, since the soft robotic arm is installed at the end of the rotating mechanism, the rotation of the mechanism will impact the dynamic end tracking of the soft robotic arm, particularly when accounting for its gravity.
To investigate this, we conduct an experiment where the rotary mechanism is commanded to move sequentially through the angles of -90$^\circ$ to -60$^\circ$, -30$^\circ$, and 0$^\circ$, as shown in Fig. \ref{fig_Exp2_2}(a).
The experimental results of the dynamic end tracking of the soft robotic arm are depicted in Fig. \ref{fig_Exp2_2}(b).
The tracking curves show that the soft robotic arm achieves precise end tracking during the movement of the rotary mechanism from -90$^\circ$ to -60$^\circ$.
When the rotary mechanism transitions to the -30$^\circ$ and 0$^\circ$ states, there exist some deviations in the end tracking in both the $x$ and $z$ directions.
This deviation may be attributed to the rapid movement of the mechanism, causing the soft robotic arm to swing, which could lead to the accumulation of particles in the stiffening layer.
Nevertheless, the deviation remains within 4.8 mm, as shown in the locally magnified image, making it acceptable for aerial manipulation tasks.

\subsection{Dynamic Transmission-Line Grasping}

\begin{figure}
	\centering
	\includegraphics[width=3.3in]{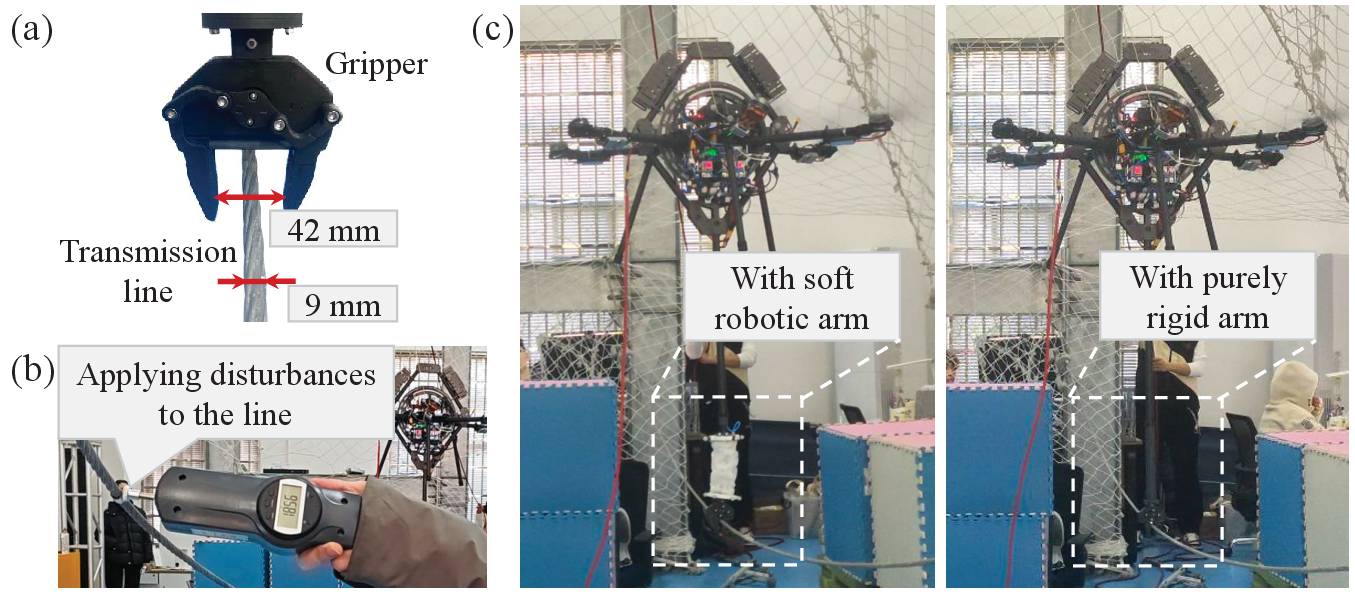}
	\caption{Setup of aerial dynamic grasping of the transmission line with comparisons.
		(a) The maximum opening size of the gripper is 42 mm and the width of the transmission line is 9 mm.
		(b) A thrust gauge is used to apply disturbances forces of different magnitudes to the line in the $x$-direction under the inertial frame $\mathcal{F}_I$.
		(c) Snapshots of the robot equipped with the soft robotic arm (left) and the purely rigid arm (right).}
	\label{fig_Exp3_Snapshot}
\end{figure}

To evaluate the compliance performance of the presented AeRSoM robot, comparative experiments are conducted involving the aerial dynamic grasping of a transmission line.
The sizes of the gripper and the line are displayed in Fig. \ref{fig_Exp3_Snapshot}(a).
During the aerial grasping phase, a thrust gauge is employed to apply various magnitudes of disturbance forces to the line in the $x$-direction, causing the grabbed line shake dynamically (Fig. \ref{fig_Exp3_Snapshot}(b)).
As depicted in Fig. \ref{fig_Exp3_Snapshot}(c), the comparative experiments assess the presented robot equipped with the soft robotic arm versus a purely rigid arm.
To ensure a fair comparison, the distance between the end of the gripper and the center of gravity of the vehicle body is kept constant, and it is assumed that the change in the overall mass of the robot caused by the replacement of the rigid arm is negligible.

\begin{figure}
	\centering
	\includegraphics[width=3.3in]{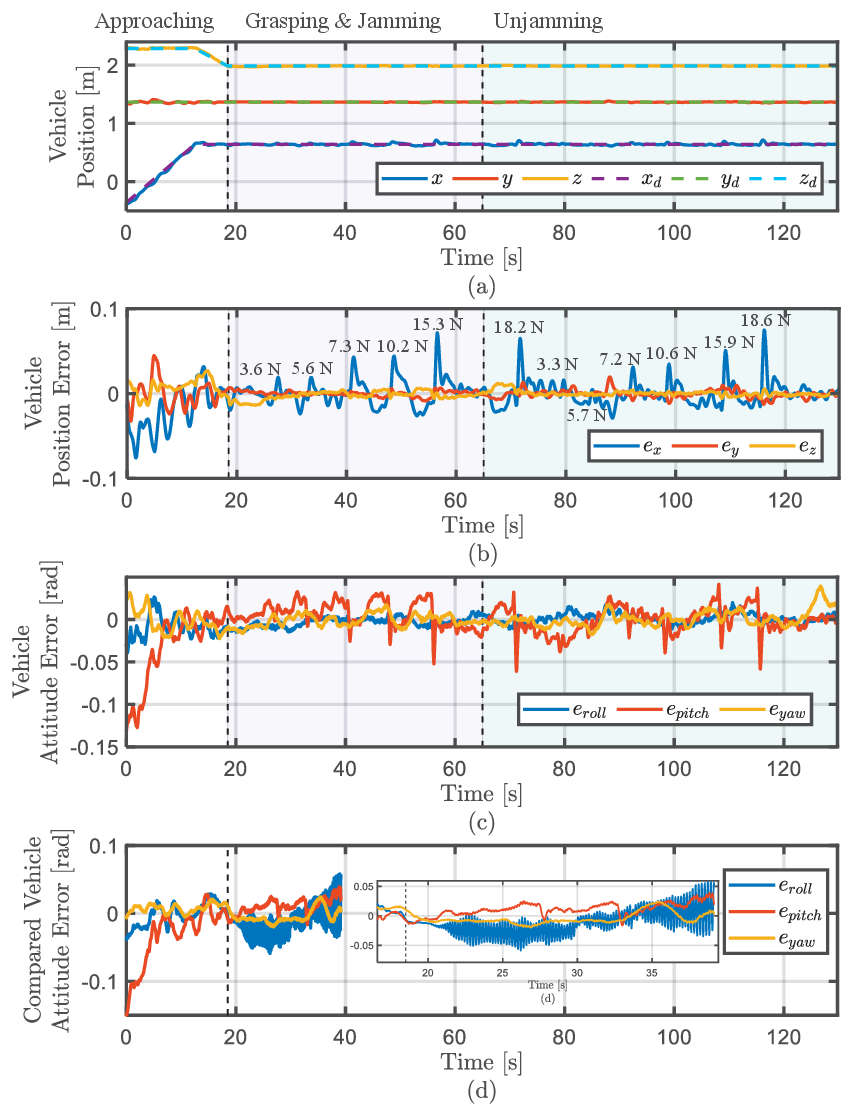}
	\caption{Comparative results of aerial grasping.
		(a)-(c) The fabricated soft robotic arm or (d) the purely rigid arm with the gripper is used for grasping.}
	\label{fig_Exp3}
\end{figure}

\begin{figure*}
	\centering
	\includegraphics[width=6.5in]{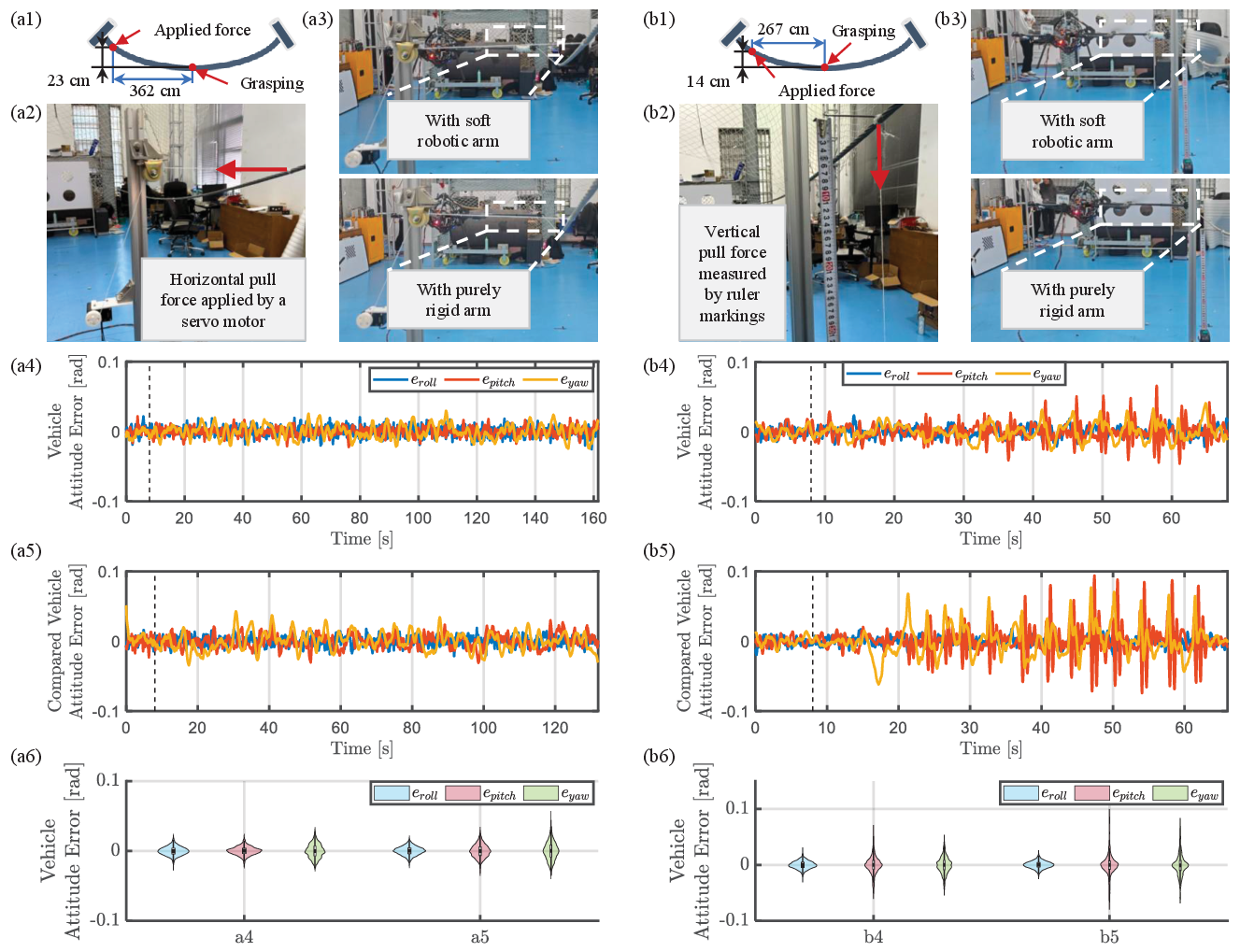}
	\caption{Comparisons of aerial dynamic horizontal grasping of the transmission line.
		(a1) and (b1) Diagrams of applied force and grasping points.
		The disturbances originate from the horizontal pull force applied by the servo motor (a2) and the vertical quantitative pull force measured by the ruler markings (b2).
		(a3) and (b3) Snapshots of the robot equipped with the soft robotic arm (top) and the purely rigid arm (bottom).
		Time trajectories of attitude errors for the soft robotic arm (a4, b4) and the purely rigid arm (a5, b5).
		Violin plots show the attitude error of the aerial vehicle under horizontal pull force (a6) and vertical pull force (b6).}
	\label{fig_Exp3_2}
\end{figure*}

Fig. \ref{fig_Exp3} presents comparative results, where the experimental results using the soft robotic arm are shown in Fig. \ref{fig_Exp3}(a)-(c).
The position trajectory tracking of the aerial vehicle is plotted in Fig.\ref{fig_Exp3}(a), which is divided into the approaching phase and the grasping phase under both the jamming and unjamming states.
Fig. \ref{fig_Exp3}(b) and Fig. \ref{fig_Exp3}(c) display the position and attitude errors of the aerial vehicle, respectively, where the applied disturbance forces are marked in Fig. \ref{fig_Exp3}(b).
Focusing on the curve of $e_x$ in Fig. \ref{fig_Exp3}(b) and taking peaks at time 56.5s and 109s as an example, a disturbance force of approximately 15 N is applied to the line while the stiffening layer of the soft robotic arm is either in the jamming state or the unjamming state.
The results indicate that the unjammed stiffening layer enables the AeRSoM robot to exhibit smaller position deviations and achieve faster convergence to the equilibrium point, thereby improving dynamic grasping stability.
Additionally, Fig. \ref{fig_Exp3}(d) plots the attitude error of the vehicle when the arm is purely rigid, and it gives that the attitude experiences severe high-frequency vibrations in the absence of a soft structure or a compliant algorithm.
This conducted experiment highlights that the presented AeRSoM robot equipped with the soft structure can safely perform aerial dynamic grasping tasks despite the dynamic movement of the grasped object, without relying on additional complex and expensive interaction methods.

To further demonstrate the performance of the proposed soft robotic arm, experimental comparisons of horizontal gripping the line are conducted using the soft arm versus the purely rigid arm, where the soft arm is bent under the effect of gravity.
Considering that the previous use of a manual thrust gauge to apply perturbations may limit the reproducibility of the results, two more controlled and reproducible validation methods are introduced.
Specifically, a servo motor with a connecting cable is used to apply a horizontal pull force through a fixed pulley (Fig. \ref{fig_Exp3_2}(a2)), and the diagram of applied force and grasping points is shown in Fig. \ref{fig_Exp3_2}(a1).
A total of ten horizontal pull forces act on the transmission line by the motor's connecting cable.
Starting from zero tension, the servo motor is commanded to rotate half a turn each time.
After each tension application, it returns to the initial zero-tension position, and this process is repeated.
Another method involves applying a manual vertical pull force measured via ruler markings (Fig. \ref{fig_Exp3_2}(b2)), where the diagram of applied force and grasping points is presented in Fig. \ref{fig_Exp3_2}(b1).
The cable is pulled vertically from the zero mark sequentially to the marks approximately 5 cm (one time), 10 cm (four times), and 16 cm (nine times).
Since the cable is still manually pulled vertically to the predetermined marks, an error of approximately 1 cm is introduced.
We hereby declare that rapidly pulling and releasing the cable in this manner can cause oscillations and interference to the end-effector.
In contrast, the servo motor, affected by torque lock, can only generate milder disturbances.
This is also the reason for constructing the second disturbance input method.
Fig. \ref{fig_Exp3_2}(a4) and Fig. \ref{fig_Exp3_2}(a5) display the experimental results of the attitude error using the soft robotic arm and the purely rigid arm under the horizontal pull force disturbances, respectively.
Fig. \ref{fig_Exp3_2}(b4) and Fig. \ref{fig_Exp3_2}(b5) plot the time trajectories of the comparative attitude error under the vertical pull force disturbances.
Violin plots visually show comparative attitude errors, as presented in Fig. \ref{fig_Exp3_2}(a6) and Fig. \ref{fig_Exp3_2}(b6).
These experiments further illustrate the compliance performance of the present AeRSoM robot equipped with the soft robotic arm, especially its ability to cope with severe oscillations.

\subsection{Physical Interaction with Wind Turbine Blade}

\begin{figure}
	\centering
	\includegraphics[width=3.3in]{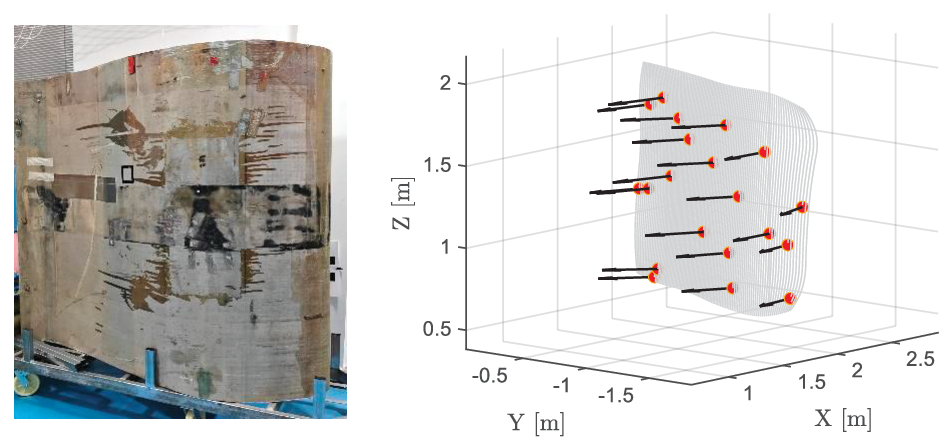}
	\caption{Wind turbine blade employed for experiments (left), and 20 random points are drawn on the surface marked with red dots and their normals are plotted as black arrows (right).}
	\label{fig_Exp4_Snapshot}
\end{figure}

\begin{figure}
	\centering
	\includegraphics[width=3.3in]{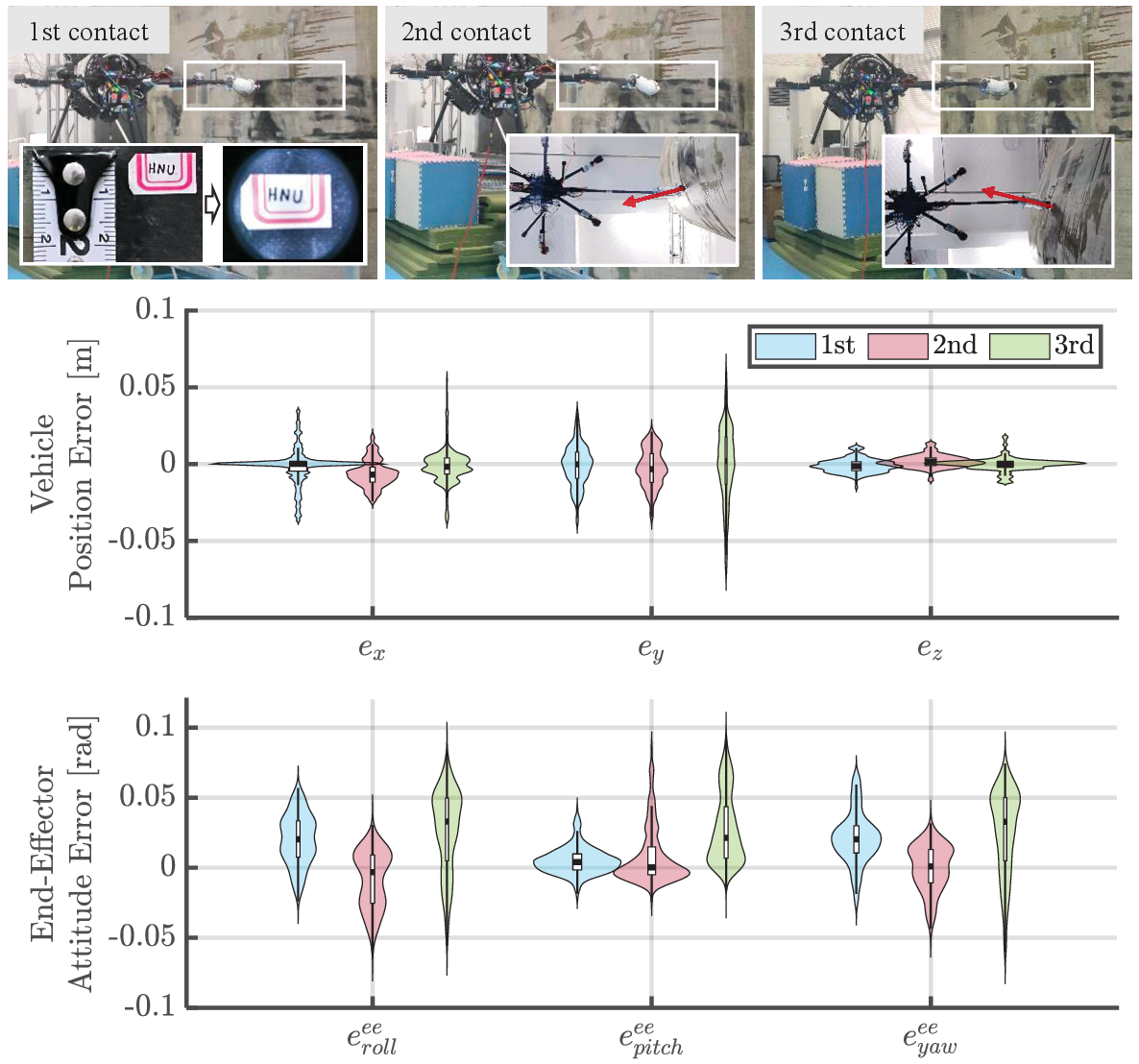}
	\caption{Snapshots of aerial physical interaction with different points on the wind turbine blade featuring an unstructured surface (top).
		Violin plots show the position tracking error of the aerial vehicle (middle) and the attitude tracking error of the end-effector (bottom) for three trials.}
	\label{fig_Exp4}
\end{figure}

To verify the flexibility of the soft robotic arm and the end-effector tracking performance, the camera probe in Fig. \ref{fig_endEffector}(b) is installed at the end of the presented AeRSoM robot to conduct aerial physical interaction with a wind turbine blade featuring an unstructured surface.
Fig. \ref{fig_Exp4_Snapshot} visualizes the contact points with their normals, and the AeRSoM robot is commanded to interact perpendicularly with the surface, where the desired contact point is selected to penetrate the surface by a constant amount to ensure that the end-effector is in contact with the surface.

The experimental results of aerial physical interaction with the unstructured surface are presented in Fig. \ref{fig_Exp4}.
The top part of Fig. \ref{fig_Exp4} displays snapshots of the end-effector in contact with three different orientations.
In the first image, a small label can be seen in the camera view, while the second and third images depict the contact in different orientations.
The middle and bottom parts of Fig. \ref{fig_Exp4} plot the position tracking error of the aerial vehicle and the attitude tracking error of the end-effector across three trials.
Since the end-effector cannot reach the desired contact point that penetrates the surface, the attitude error of the end-effector is considered as an important evaluation metric, which is affected by the position and attitude errors of both the vehicle body and the soft robotic arm.
The position tracking error of the vehicle during the third contact is slightly larger compared to the first two contacts, indicating a relatively larger attitude error for the end-effector in that instance.
Throughout the entire process, the stiffening layer of the soft robotic arm remains jammed.
In the free flight phase, the soft robotic arm is stiffened to mitigate end-effector vibrations.
Our tests reveal that if the stiffening layer is in an unjammed state during contact, the soft robotic arm becomes vulnerable to significant shearing effects.

\subsection{Peg-in-Hole and Screwing}

\begin{figure}
	\centering
	\includegraphics[width=3.3in]{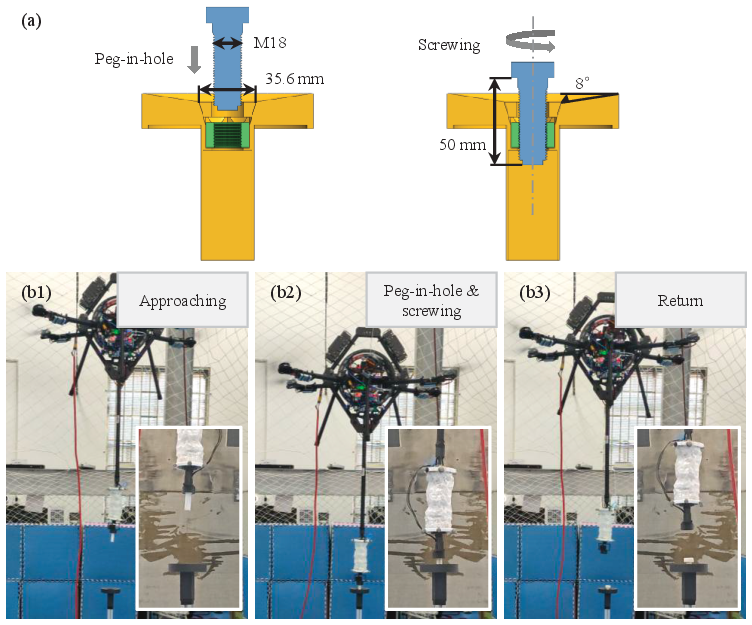}
	\caption{Setup of aerial peg-in-hole and screwing task (a) and key snapshots during this manipulation, including approaching (b1), aerial peg-in-hole and screwing (b1), and mission completion and return (b3).}
	\label{fig_Exp5_Snapshot}
\end{figure}

\begin{figure*}
	\centering
	\includegraphics[width=6.9in]{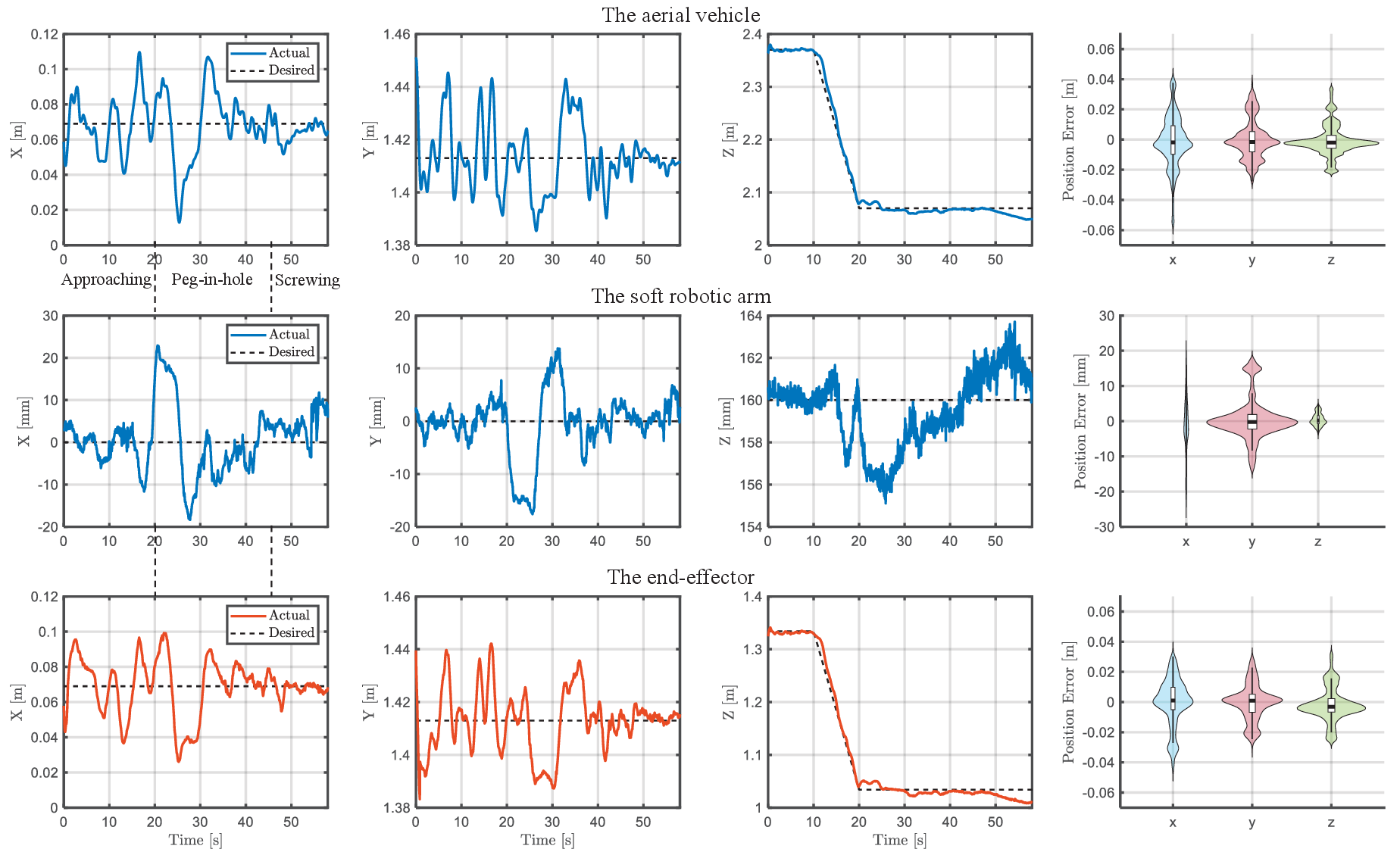}
	\caption{Trajectory of the AeRSoM during aerial peg-in-hole and screwing a bolt.}
	\label{fig_Exp5}
\end{figure*}

To further investigate the capability of the AeRSoM robot to perform precise manipulation tasks, a challenging aerial bolt-tightening task is developed in this subsection.
The procedure starts with the placement of the bolt into a peg-in-hole configuration, followed by the activation of the end-effector illustrated in Fig. \ref{fig_endEffector}(c) to tighten the bolt, as shown in Fig. \ref{fig_Exp5_Snapshot}(a).
Given that the bolt and nut must fit together with minimal tolerance, this manipulation task could become quite troublesome.
In this experiment, a M18$\ast$50 mm bolt is fixed at the end-effector and inserted into a hole with a diameter of 35.6 mm, where the upper surface of the hole is inclined at approximately 8$^\circ$, allowing the bolt to slide into the hole.

Fig. \ref{fig_Exp5_Snapshot} presents the key snapshots from the bolt-tightening operation captured in the video, highlighting the following processes:
1) the aerial vehicle makes an effort to carry out the bolt to approach the hole (Fig. \ref{fig_Exp5_Snapshot} (b1);
2) the aerial vehicle collaborates with the soft robotic arm to insert the bolt from the end-effector into the hole allowing a play of $\pm$8.8 mm, and then the bolt is tightened into the nut (Fig. \ref{fig_Exp5_Snapshot} (b2);
3) the bolt-tightening mission is completed, and the vehicle returns (Fig. \ref{fig_Exp5_Snapshot} (b3).
Fig. \ref{fig_Exp5} plots the trajectory tracking performance of the aerial vehicle, soft robotic arm and end-effector during aerial peg-in-hole and screwing tasks.
During the free flight approach and insertion phases, the stiffening layer of the soft robotic arm is jammed to increase the end rigidity, allowing for the precise insertion of the bolt into the hole.
Once the tightening operation begins, the stiffening layer is unjammed to provide sufficient compliance and safety.
Three sets of successful experiments shown in the video are conducted to confirm that the completion of accurate peg-in-hole and screwing tasks is not merely accidental.
To further clarify the necessity of variable stiffness of the soft robotic arm, a comparative experiment is conducted, as exhibited in the video, where the soft robotic arm remains in the unjammed state throughout.
Experimental observations reveal that the unjammed soft robotic arm results in low end-effector accuracy, making it difficult to achieve the hole-docking task.

\section{Discussions}

The experimental results demonstrate that the performance improvement of the proposed AeRSoM is not solely attributed to the control framework, but also to the physical morphology of the system.
This distributed embodied compliance enhances disturbance tolerance during contact-rich manipulation while reducing the stabilization burden on the flight system.

Another key observation is that aerial manipulation inherently involves a tradeoff between compliance and precision.
While Highly compliant manipulators improve interaction safety and robustness, they often suffer from reduced positioning accuracy and force transmission capability.
In contrast, rigid manipulators enable precise manipulation but amplify contact disturbances and increase the risk of destabilizing the aerial platform.
The proposed rigid-soft architecture seeks to balance these competing requirements by combining passive compliance with task-oriented structural rigidity.
Furthermore, the variable-stiffness mechanism allows the manipulator to adapt its mechanical properties according to different manipulation stages, thereby improving both interaction robustness and task execution performance.
The experimental results indicate that stiffness modulation is particularly beneficial for tasks involving both environmental uncertainty and high precision requirements, such as peg-in-hole and screwing operations.

More broadly, the proposed system highlights the potential of embodied intelligence for aerial manipulation.
Rather than relying solely on increasingly complex sensing and control algorithms, part of the interaction complexity can be handled through physical morphology and mechanical intelligence.
The distributed compliance of the rigid-soft manipulator allows the robot to passively adapt to uncertain contacts, thereby reducing the burden on perception, force estimation, and active stabilization.
This observation suggests that integrating morphology, materials, and control may provide an effective pathway toward more capable aerial manipulation systems.

Despite the encouraging results, several limitations remain.
The experiments are conducted in an indoor laboratory setting, where state feedback is provided by a motion capture system.
If the AeRSoM robot is to be used outdoors, onboard sensing will be required both for the aerial vehicle and the soft robotic arm.
In particular, the aircraft can utilize cameras, GPS, or LIDAR for outdoor positioning.
Regarding the soft robotic arm, soft stretch sensors can be arranged along the main axis of the soft robotic arm to enable distributed bending measurement \cite{2023HuangSciAdv}.
The chamber pressure sensor is installed at the air supply pipeline, providing a rapid dynamic response and a measurement range that fully covers the operating pressure.
Further, the optical fiber sensing offers excellent electromagnetic tolerance and supports high-density distributed strain measurement, making it suitable for high-voltage or harsh environments, such as areas near power lines \cite{2019GallowaySoftRobotics}.
Moreover, the IMU sensor, mounted at the end-effector or other critical rigid connections, can provide essential information on the end-effector's pose and dynamic behavior \cite{2023PengTSMC}.
Therefore, integrating soft stretch sensors, pressure sensors, and IMU sensors presents a promising approach to accurately capture both the pose and dynamic states of the proposed soft robotic arm, facilitating its practical applications.
For high-voltage or other harsh environments, fiber optic sensing can be considered as a suitable alternative to soft stretch sensors.
In terms of power supply, a 16000mAh-6S-25C LiPo battery is employed.
When the aircraft and the flexible robotic arm operate simultaneously, the presented robot can operate for approximately 8.4 minutes while the voltage drops from 25.0 V to 22.5 V when the aircraft and the flexible robotic arm operate simultaneously, allowing the presented robot to operate for approximately 8.4 minutes.
Alternatively, another feasible solution is to use tethered power.

\section{Conclusion}

This article presented AeRSoM, a novel aerial rigid-soft integrated manipulator designed for contact-rich aerial manipulation.
The proposed system combines a fully actuated aerial platform, a rigid-soft manipulator, and a variable-stiffness mechanism to achieve stable flight, compliant interaction, and precise task execution.
A composite control framework was further developed to improve end-effector tracking performance during aerial manipulation.
Extensive experiments demonstrated the effectiveness of the proposed system in diverse contact-rich scenarios.
The results show that distributed embodied compliance can improve interaction robustness while preserving sufficient manipulation accuracy for task execution.
Overall, this work suggests that rigid-soft integration provides a promising approach to balancing the tradeoff between interaction compliance and manipulation precision in aerial robots.
Future work will focus on force-aware interaction control, faster stiffness modulation, and more autonomous aerial manipulation in complex environments.



\ifCLASSOPTIONcaptionsoff
  \newpage
\fi

\begin{IEEEbiography}[{\includegraphics[width=1in,height=1.25in,clip,keepaspectratio]{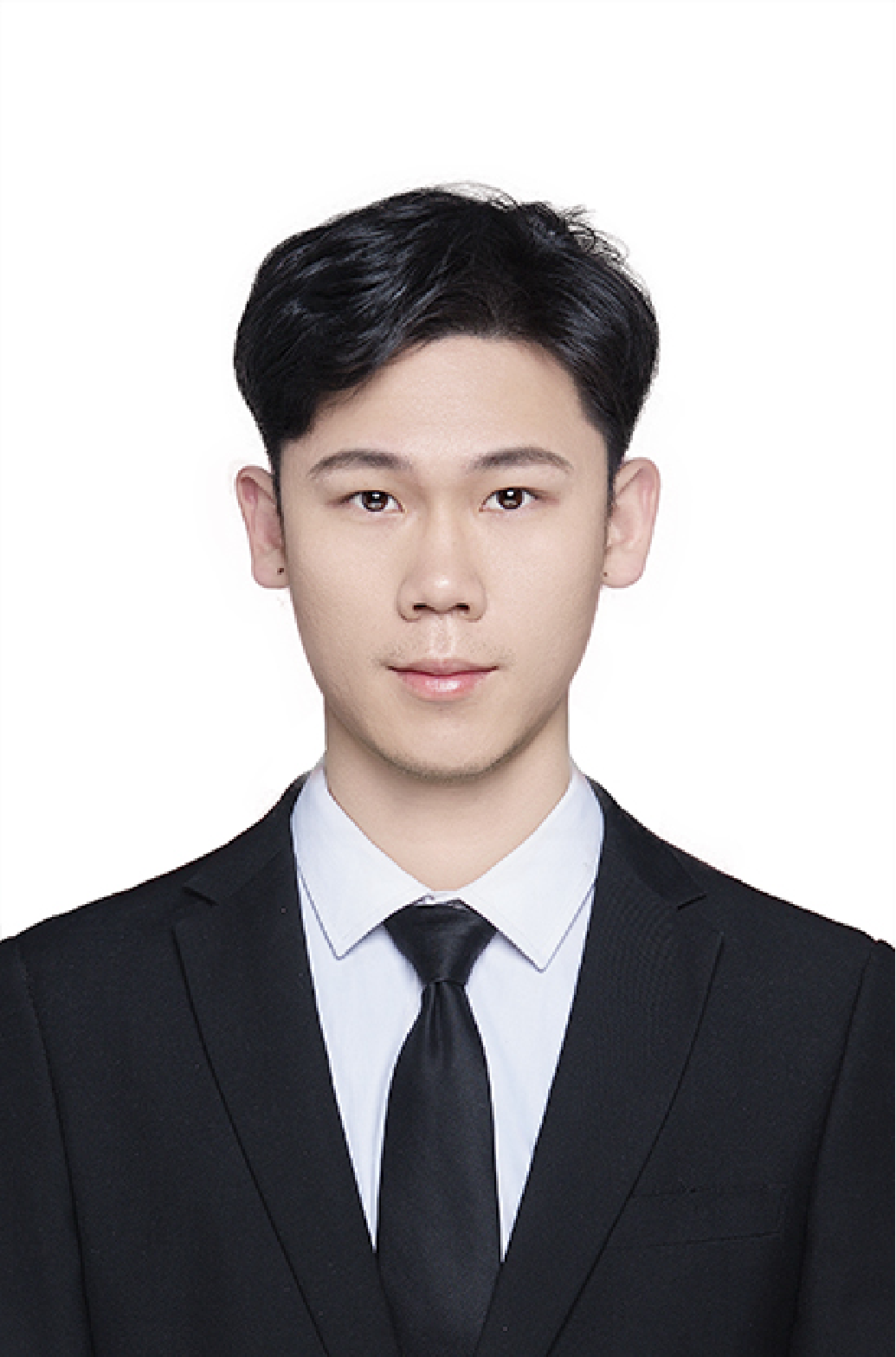}}]{Jiacheng Liang}
received the B.S. degree in mechanical design manufacture and automation and the M.S. degree in mechatronic engineering from Fuzhou University, Fuzhou, China, in 2019 and 2022, and the Ph.D. degree in control science and engineering from Hunan University, Changsha, China, in 2026.
He is currently a Postdoctoral Research Fellow with the Hunan University.
His research interests include aerial robotics, soft robotics, robot control, and aerial manipulation.
\end{IEEEbiography}

\begin{IEEEbiography}[{\includegraphics[width=1in,height=1.25in,clip,keepaspectratio]{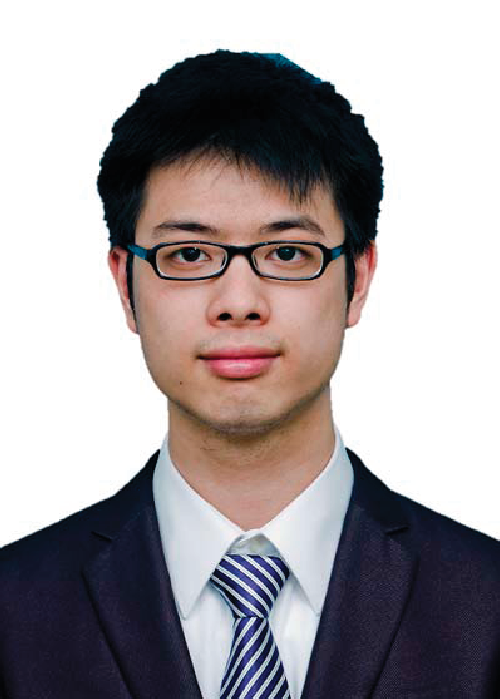}}]{Hang Zhong}
(Member, IEEE) received the B.S., M.S., and Ph.D. degrees in automation science from the College of Electrical and Information Engineering, Hunan University, Changsha, China, in 2013, 2016, and 2020 respectively.
From 2020 to 2022, he was a post-doc fellow with the Department of Electrical and Information Engineering, Hunan University, Changsha, China.
He is currently an Associate Professor with the School of Robotics, Hunan University, Changsha, China.
His research interests include aerial robotics, multi-robot systems, visual servoing, visual navigation and nonlinear control.
\end{IEEEbiography}

\begin{IEEEbiography}[{\includegraphics[width=1in,height=1.25in,clip,keepaspectratio]{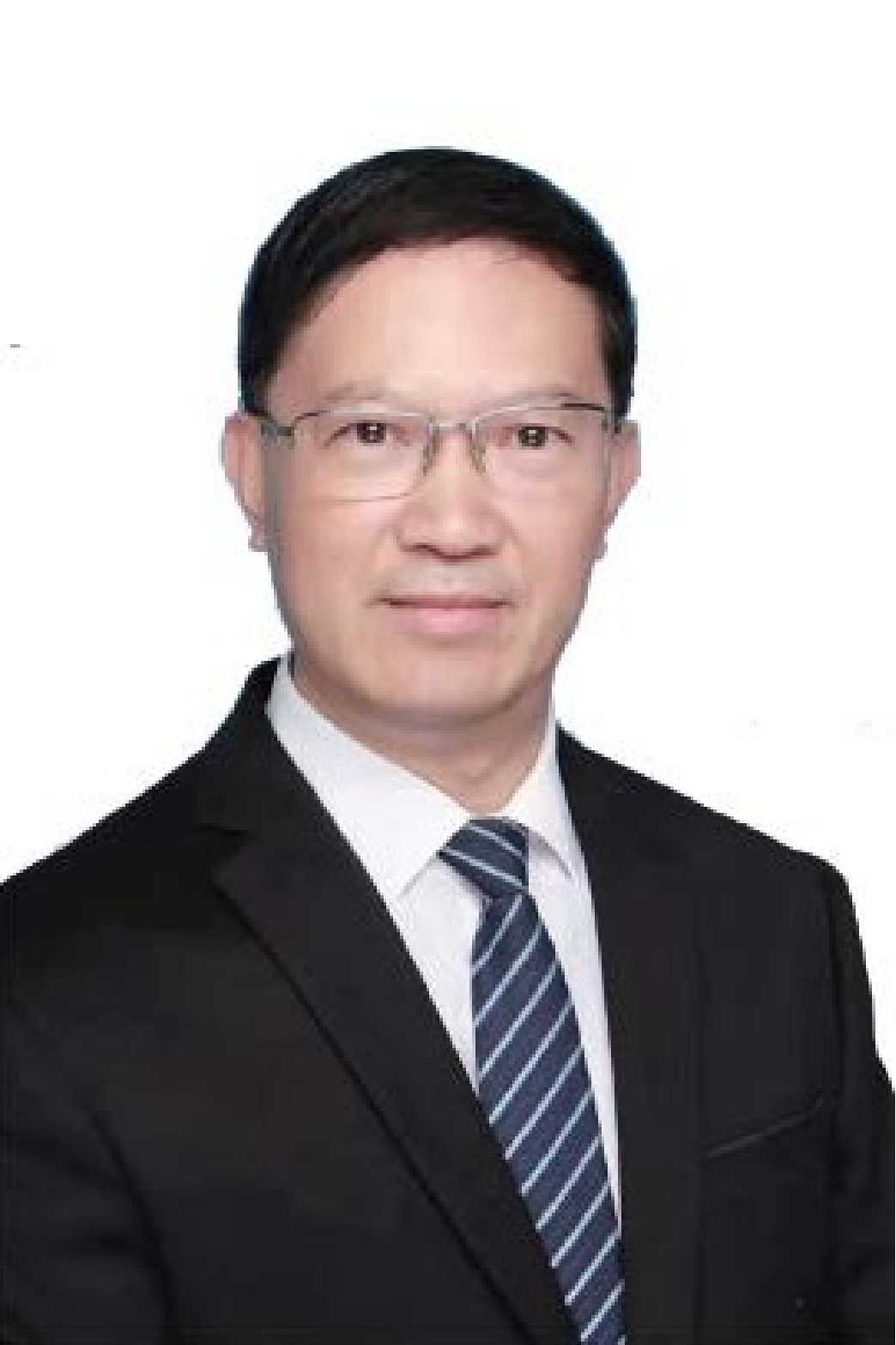}}]{Yaonan Wang}
received the B.S. degree in computer engineering from East China University of Science and Technology, Fuzhou, China, in 1981 and the M.S. and Ph.D. degrees in control engineering from Hunan University, Changsha, China, in 1990 and 1994, respectively.
He was a Post-Doctoral Research Fellow with the National University of Defense Technology, Changsha, from 1994 to 1995, a Senior Humboldt Fellow in Germany from 1998 to 2000, and a Visiting Professor with the University of Bremen, Bremen, Germany, from 2001 to 2004.
He has been a Professor with Hunan University since 1995.
His research interests include robot control, intelligent control and information processing, industrial process control, and image processing.
He has been an academician of China Engineering Academy since 2019.
\end{IEEEbiography}

\begin{IEEEbiography}[{\includegraphics[width=1in,height=1.25in,clip,keepaspectratio]{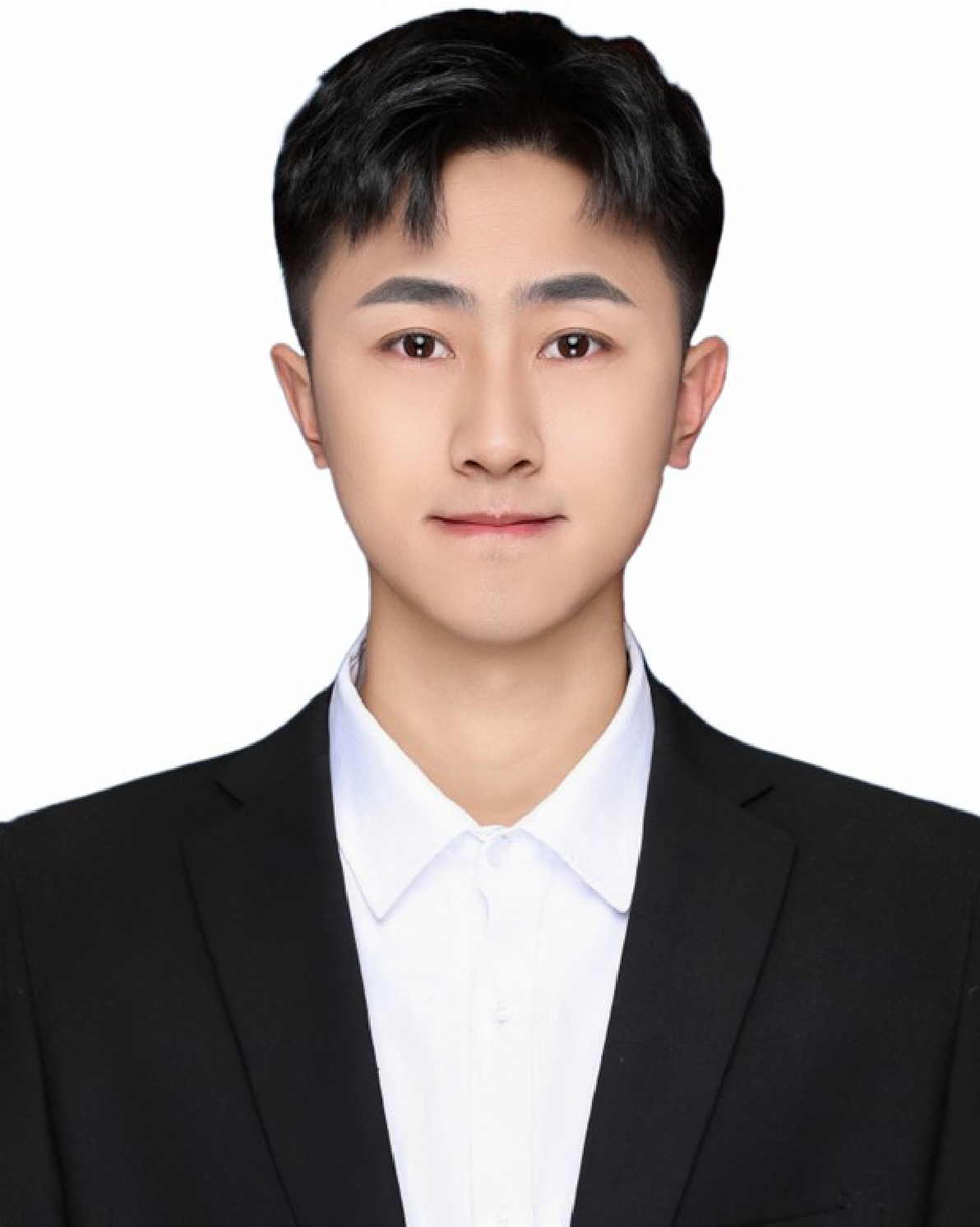}}]{Ge Chen}
received the B.S. degree in artificial intelligence from China University of Mining and Technology in 2020.
He is currently working toward the master's degree in control science and engineering at Hunan University, Changsha, China.
His research interests include aerial robotics and robot control.
\end{IEEEbiography}

\begin{IEEEbiography}[{\includegraphics[width=1in,height=1.25in,clip,keepaspectratio]{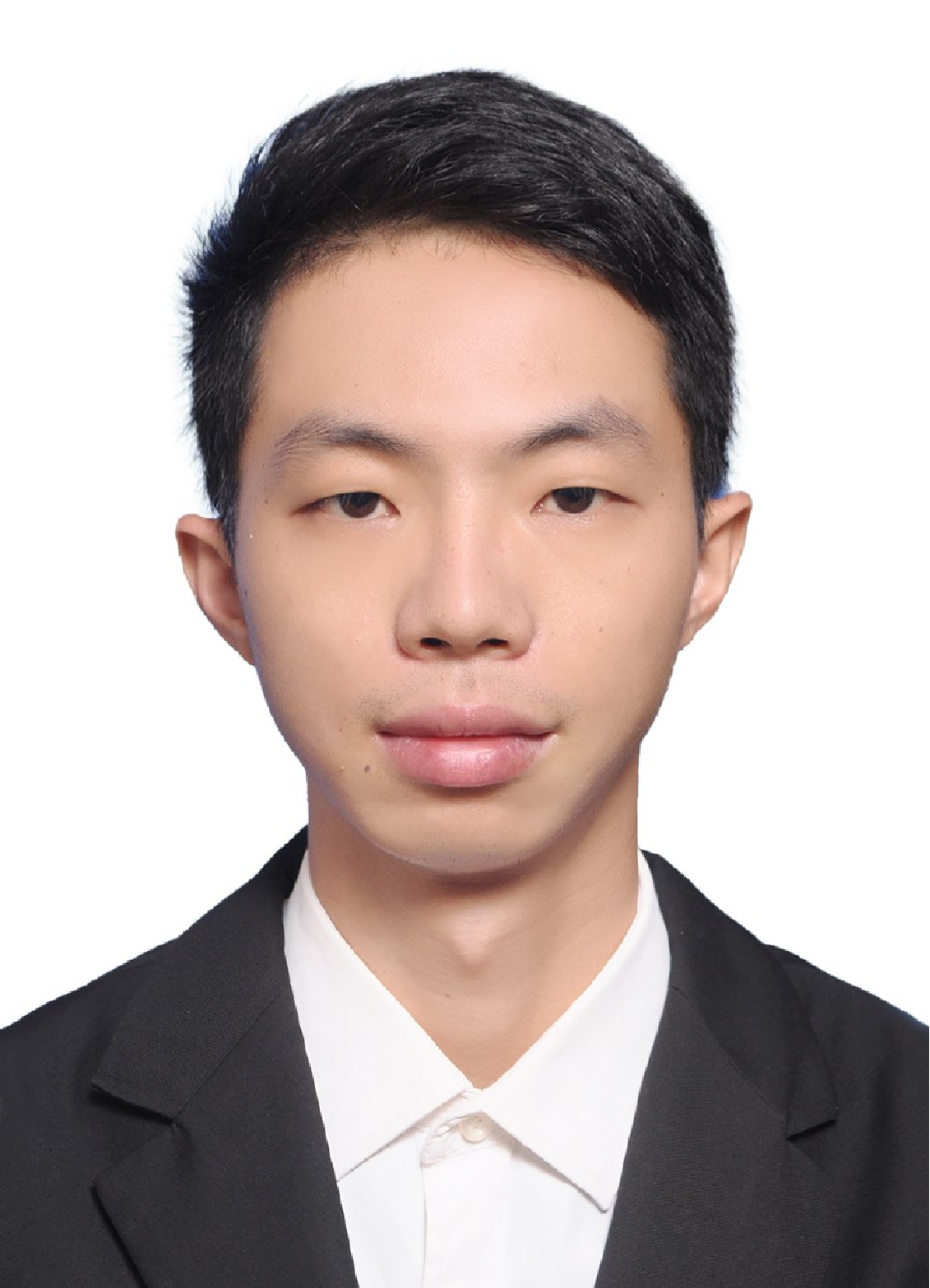}}]{Zhixing Zhang}
received the B.S.degree in mechanical design manufacture and automation from Guangdong University of Technology, Guangzhou, China in 2021, the M.S.degree in mechatronic engineering from Fuzhou University, Fuzhou, China in 2024.
He is currently working toward the Ph.D. degree in control science and engineering from Hunan University, Changsha, China.
His research interests include motion planning and mobile robot.
\end{IEEEbiography}

\begin{IEEEbiography}[{\includegraphics[width=1in,height=1.25in,clip,keepaspectratio]{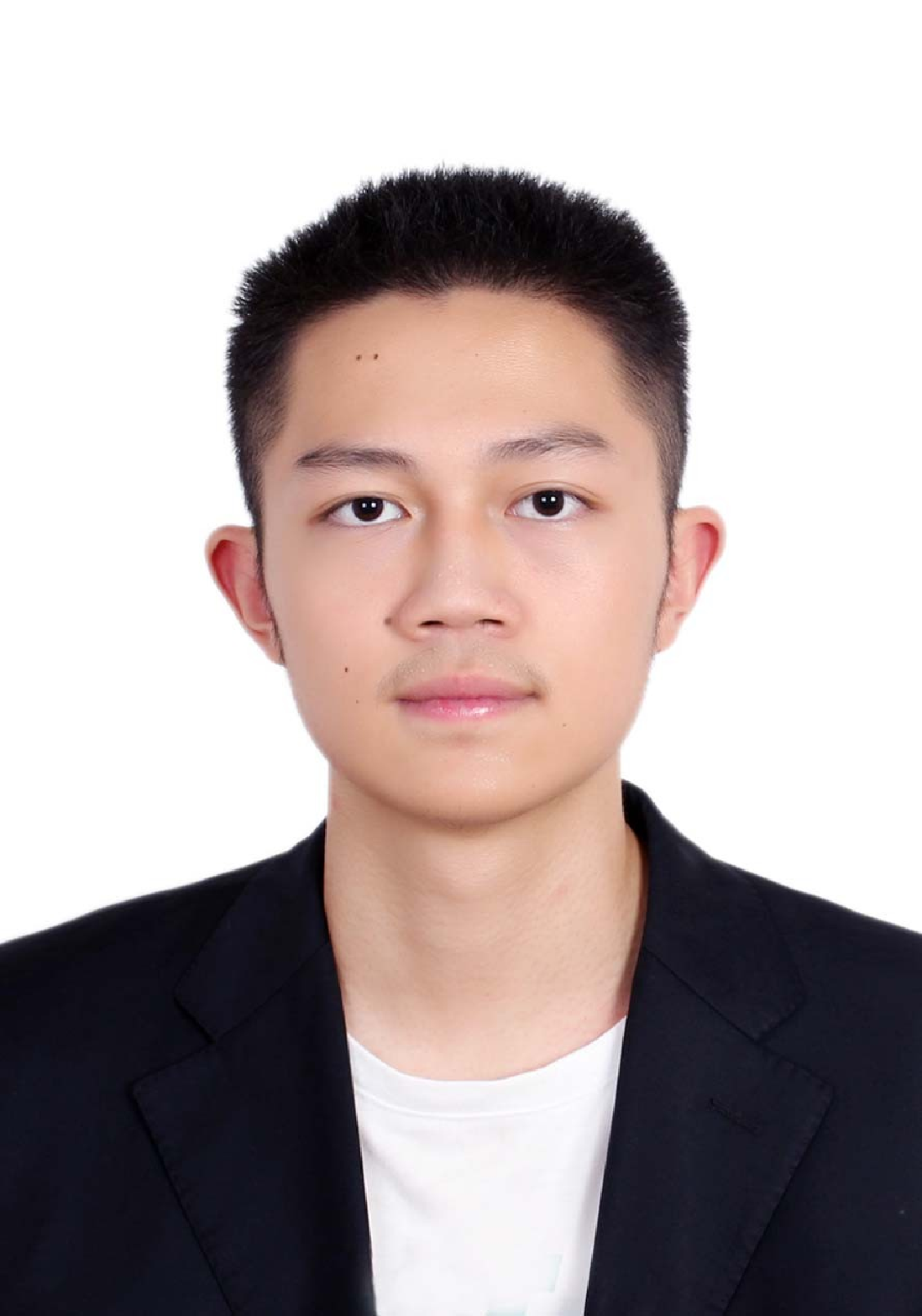}}]{Bocheng Tian}
received the B.S. degree in engineering mechanics in 2023 from Beihang University, Beijing, China, where he is currently working toward the Ph.D. degree in mechanical engineering.
His research interests include design and control of aerial and aquatic robots, soft robots, and biomimetic adsorption mechanism.
\end{IEEEbiography}

\begin{IEEEbiography}[{\includegraphics[width=1in,height=1.25in,clip,keepaspectratio]{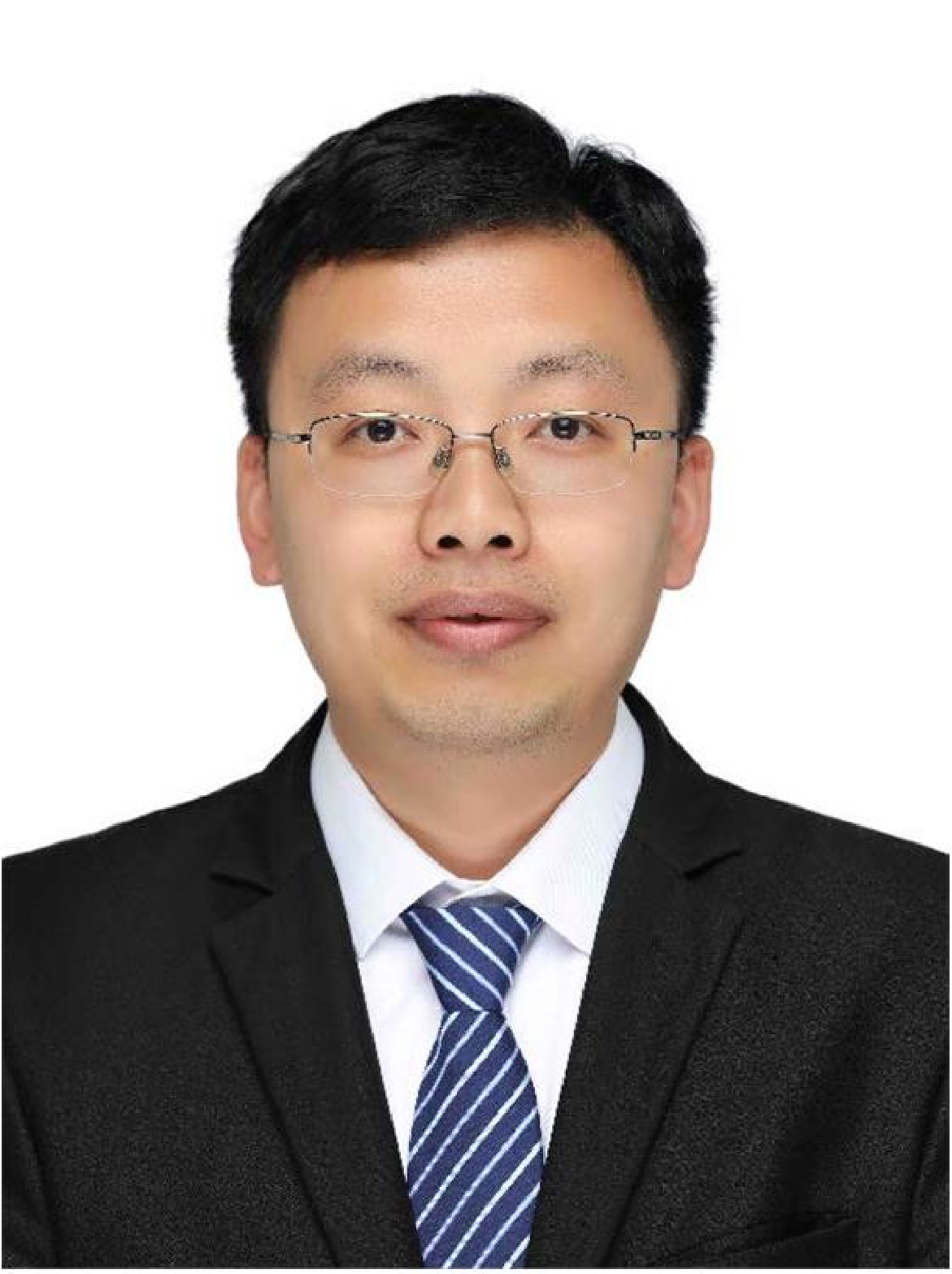}}]{Hui Zhang}
(Member, IEEE) received the B.S., M.S., and Ph.D. Degrees in pattern recognition and intelligent system from Hunan University, Changsha, China, in 2004, 2007, and 2012, respectively.
He is currently a professor with the School of Robotics, Hunan University, and he is the Deputy Director of the National Engineering Research Center of Robotic Vision Perception and Control Technology.
He was a Visiting Scholar with Common Vulnerability Scoring System Laboratory, Department of Electrical and Computer Engineering, University of Windsor, Windsor, ON, Canada, in 2017.
His research interests include machine vision, sparse representation, and visual tracking.
\end{IEEEbiography}

\begin{IEEEbiography}[{\includegraphics[width=1in,height=1.25in,clip,keepaspectratio]{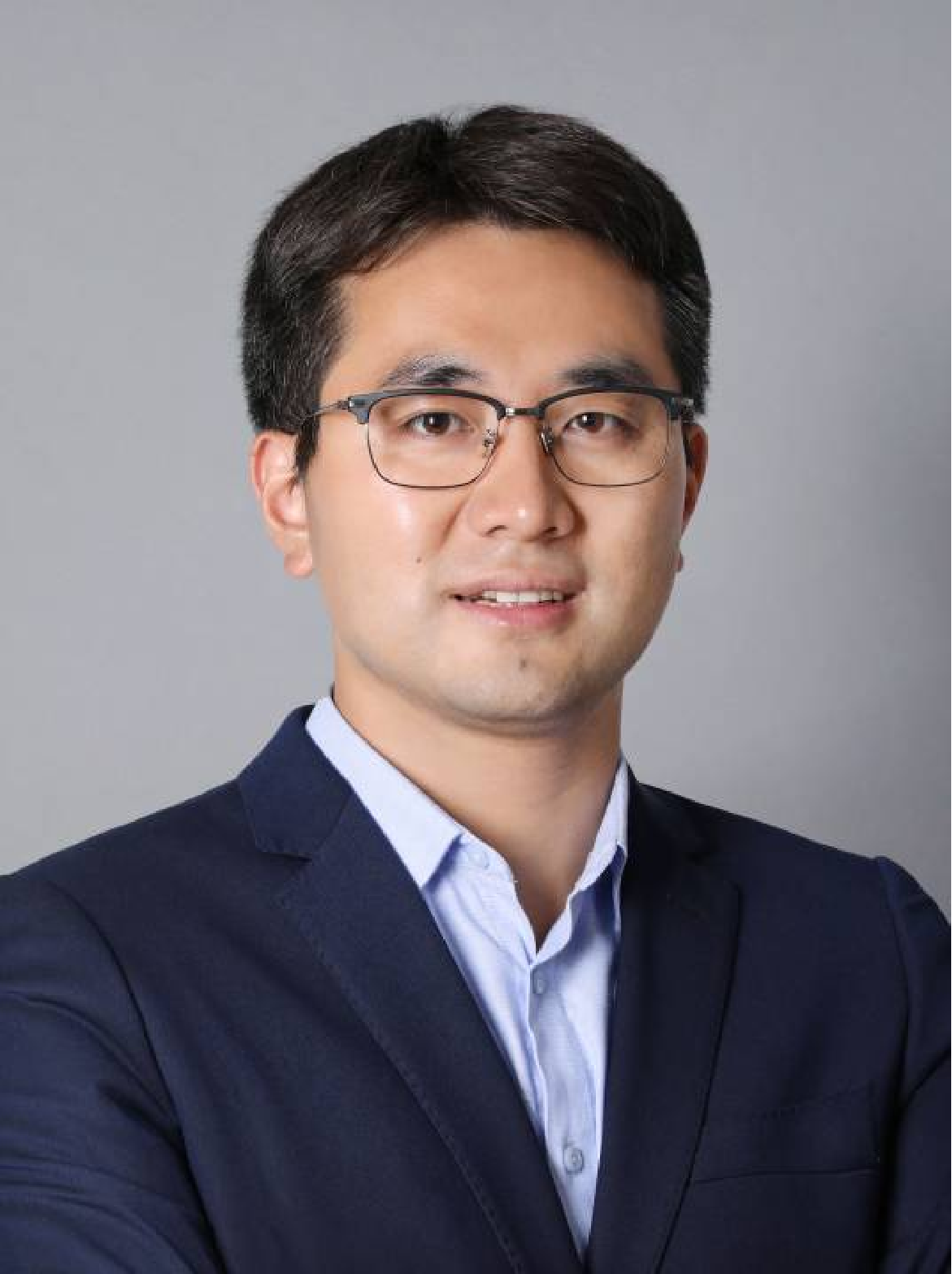}}]{Li Wen}
(Member, IEEE) received the bachelor's degree from the Beijing Institute of Technology, Beijing, China, in 2005, and the Ph.D. degree
from Beihang University, Beijing, in 2011.
He is currently a Full Professor with the Department of Mechanical Engineering, Beihang University.
He is the Vice Dean of the Department of Mechanical Engineering, Beihang University.
His current research interests include soft robots, bioinspired robotics, and embodied intelligence for robots.
Dr. Wen is an Associate Editor of IEEE Transactions on Robotics, International Journal of Robotics Research, Soft Robotics, IEEE Robotics and Automation Letters, etc.
\end{IEEEbiography}

\end{document}